\documentclass{article}

\usepackage{arxiv}

\usepackage[utf8]{inputenc} % allow utf-8 input
\usepackage[T1]{fontenc}    % use 8-bit T1 fonts
\usepackage{hyperref}       % hyperlinks
\usepackage{url}            % simple URL typesetting
\usepackage{booktabs}       % professional-quality tables
\usepackage{amsfonts}       % blackboard math symbols
\usepackage{nicefrac}       % compact symbols for 1/2, etc.
\usepackage{microtype}      % microtypography
\usepackage{lipsum}		% Can be removed after putting your text content
\usepackage{graphicx}
\usepackage{natbib}
\usepackage{doi}

\usepackage{soul}
\usepackage[utf8]{inputenc}
\usepackage{graphicx}
\usepackage{amsmath}
\usepackage{amsthm}
\usepackage{booktabs}

\usepackage{epsfig}

\usepackage{amssymb}
\usepackage{epsfig}
\usepackage{mathtools}
\usepackage[boxruled]{algorithm2e}

\usepackage{multirow}
\usepackage{booktabs}
\usepackage{balance}
\usepackage{subfig}

\newtheorem{theorem}{Theorem}
\newtheorem{definition}{Definition}
\newtheorem{lemma}{Lemma}

\DeclareMathOperator{\Tr}{Tr}

\usepackage[colorinlistoftodos]{todonotes}

\title{Asymptotic Causal Inference}

\date{} 					% Or removing it

\author{
       Sridhar Mahadevan\\
   Adobe Research, 345 Park Avenue, San Jose, CA 95110 \\
   
   smahadev@adobe.com
}

\begin{document}

\maketitle

\begin{abstract}
We investigate causal inference in the asymptotic regime as the number of variables $n \rightarrow \infty$ using an information-theoretic framework. We define structural entropy of a causal model in terms of its description complexity measured by the logarithmic growth rate, measured in bits, of all directed acyclic graphs (DAGs) on $n$ variables, parameterized by the edge density $d$. Structural entropy yields non-intuitive predictions. If we randomly sample a DAG  from the space of {\em all} models over $n$ variables, as $n \rightarrow \infty$,  in the range $d \in (0, \frac{1}{8})$, almost surely ${\cal D} $ is a two-layer DAG!  Semantic entropy quantifies the reduction in entropy where edges are removed by causal intervention.  Semantic causal entropy is defined as the $\phi$-divergence $D_{\phi}(P \parallel P_S)$  between the observational distribution $P$ and the interventional distribution $P_S$, where a subset $S$ of edges are intervened on to determine their causal influence. We compare the decomposability properties of semantic entropy for different choices of $\phi$, including $\phi(t) = t \log t$ (KL-divergence),  $\phi = \frac{1}{2}(\sqrt{t} - 1)^2$ (squared Hellinger distance), and $\phi = \frac{1}{2}|t-1|$ (total variation distance). We apply our framework to generalize a recently popular bipartite experimental design for studying causal inference on large datasets, where interventions are carried out on one set of variables (e.g., power plants, items in an online store), but outcomes are measured on a disjoint set of variables (residents near power plants, or shoppers). We generalize bipartite designs to $k$-partite designs, and describe an optimization framework for finding the optimal $k$-level DAG architecture for any value of $d \in (0, \frac{1}{2})$. As $d$ increases, a sequence of phase transitions  occur over disjoint intervals of $d$, with deeper DAG architectures emerging as $d \rightarrow \frac{1}{2}$.  We also give a quantitative bound on the number of samples needed to reliably test for average causal influence for a $k$-partite design.
\end{abstract}
\section{Introduction}
\begin{figure}[t]
\centering
\begin{minipage}{0.45\textwidth}
\centering
\vspace{0pt}
\includegraphics[scale=0.17]{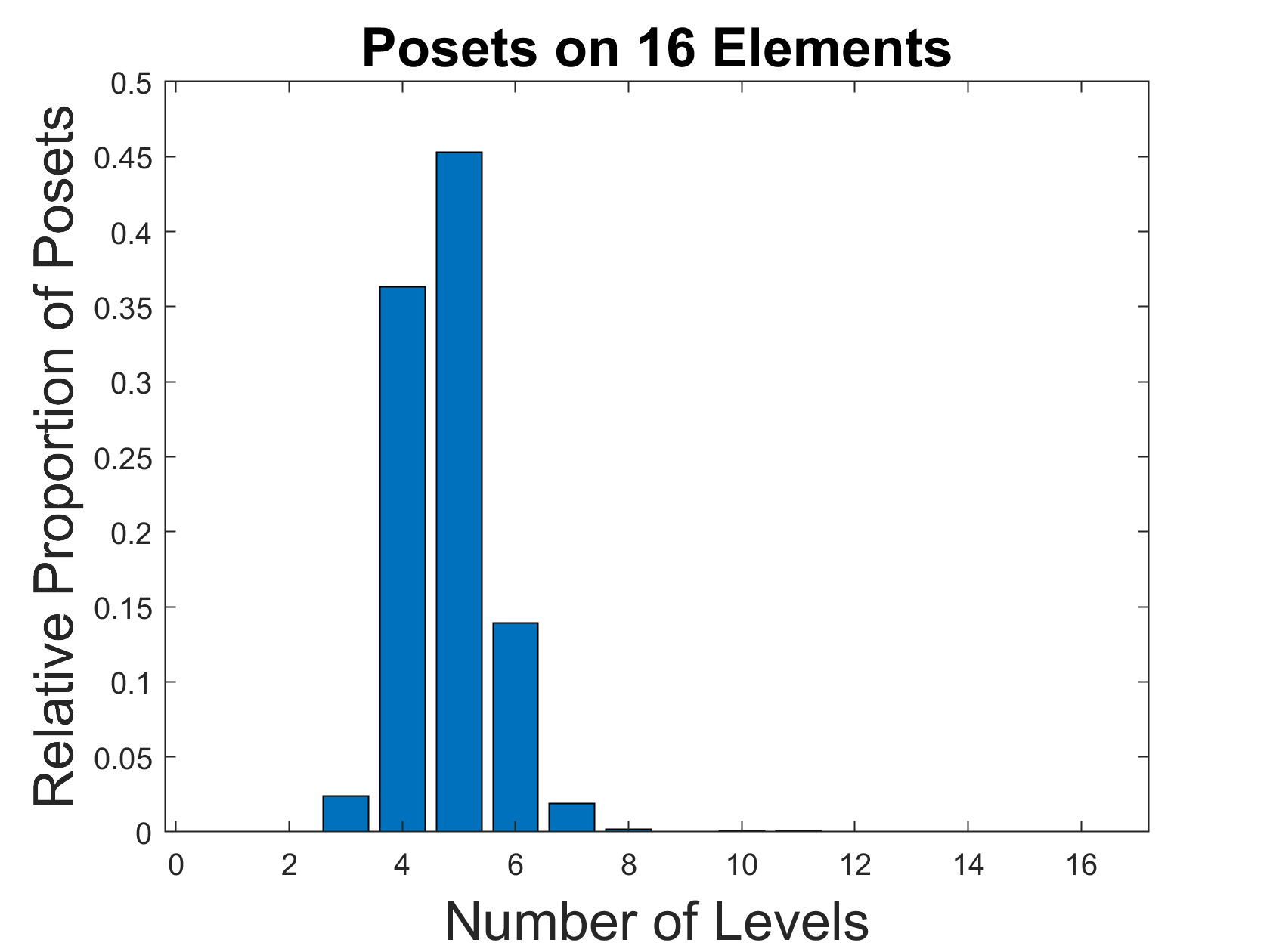}
\end{minipage} \hfill
\begin{minipage}{0.45\textwidth}
\centering
\vspace{0pt}
\includegraphics[scale=0.17]{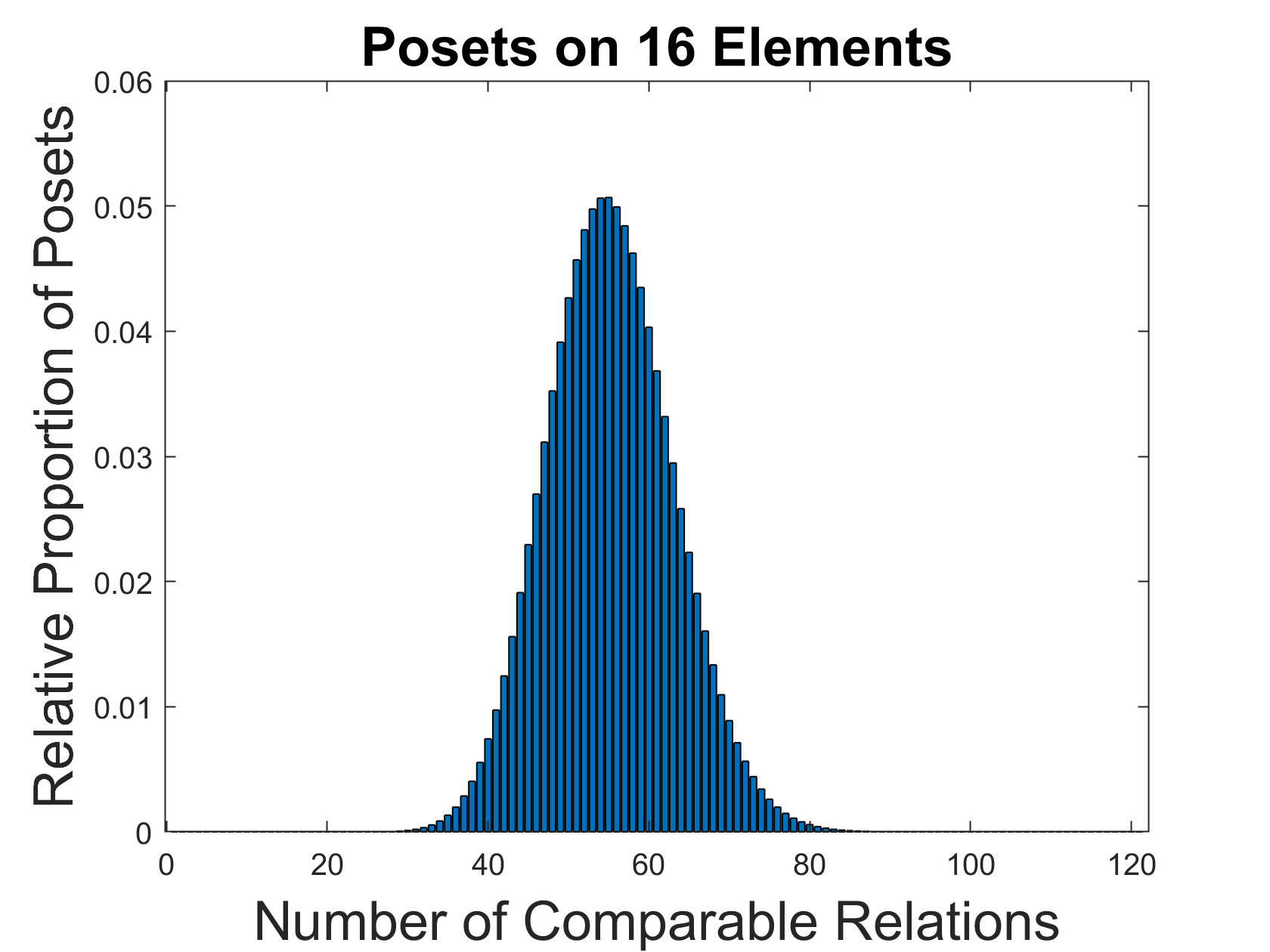}
\end{minipage}
\caption{The relative proportion of non-isomorphic posets (or DAGs) on $16$ elements as a function of the number of levels (left) and the number of relations or edges (right). Note the highly peaked distribution: randomly sampling the space of $16$-variable models will with high probability generate a DAG of $4-6$ levels with $50-70$ comparable relations.  Plots based on tables in \citep{brinkmann}.}
%Structural entropy measures asymptotic model complexity, as a function of the percentage $d \in (0,1)$ of possible relations $\mathcal{O}(n^2)$ in a poset. $d=0$ yields a single layer decoupled poset with one antichain $\mathcal{O}(n)$ set of incomparable elements. $d = \frac{1}{2}$ is exemplified by an $n$ layer Markov chain poset with ${n \choose 2}$ comparable elements, whose antichain sets are singletons. $k$-layer models of intermediate complexity are marked by phase transition regions of $d$. Semantic entropy is defined using {\em directed channel capacity}, a measure of the asymptotic rate of information transmission in causal models by interventions.} 
\label{diagram1}
\end{figure}
Inspired by a range of asymptotic studies,  from neural tangent kernels \citep{DBLP:conf/nips/JacotHG18} to random graphs \citep{DBLP:journals/siamdm/FriezeT20} and phase transition effects in satisfiability problems \citep{DBLP:journals/dam/BaileyDK07},
we investigate causal inference in a novel regime as the number of variables $n \rightarrow \infty$. In contrast, most previous work that has investigated the non-asymptotic case \citep{pearl:causalitybook,spirtes:book,DBLP:conf/uai/Eberhardt08,DBLP:journals/jmlr/HauserB12,DBLP:conf/nips/KocaogluSB17,MAOCHENG198415,prasad:aaai21,pmlr-v65-daskalakis17a}. Our approach is motivated by the need to scale causal inference to real-world applications in diverse areas such as improving healthcare outcomes by reducing pollution, social network analysis, recommender systems, online ad placement, and two-sided platforms for dynamic pricing, which may involve building models over potentially millions of variables \citep{DBLP:journals/geoinformatica/LiFZMTZ20,DBLP:conf/nips/Pouget-AbadieAS19,DBLP:conf/kdd/Schlosser018,DBLP:conf/ijcai/SchlosserW0U18,DBLP:journals/entropy/RodderDKLR19,DBLP:conf/sigecom/CharlesCDJS10,zigler2018bipartite}. We define an information-theoretic framework based on an evolutionary process of growth and decay of the relative proportion $O(d n^2)$ of edges (or relations) in the causal model. Structural causal entropy, or model description complexity, quantifies the evolutionary growth in the number of models, where we build on some classic results in extremal combinatorics of partially ordered sets (posets) \citep{dhar:1978,kleitman:1979,promel:phase}.  Semantic entropy, in contrast, quantifies the reverse evolutionary process of the decay in the relative proportion of edges due to causal intervention.   In particular, we build on the information-theoretic paradigm  for quantifying causal influence  \citep{DBLP:conf/isit/MasseyM05,wieczorek,DBLP:conf/allerton/Raginsky11,nihat,janzing}. We use the {\em edge-centric} paradigm of causal intervention proposed in \citep{janzing}, except we generalize their approach to using a general $\phi$-divergence. 

If we imagine causal discovery as nature generating data from a randomly chosen causal model from the space of all possible causal DAG models on $n$ variables, as $n \rightarrow \infty$, surprisingly, almost surely the DAG has a very small number of levels (see Figure~\ref{diagram1}).    \citep{kleitman:1975} derived upper bounds on the logarithmic growth rate of DAGs and their associated posets ${\cal P}_n$, as $n \rightarrow \infty$. Their analysis showed that that the class of all $n$-layer DAG models ${\cal P}_n$ is dominated in the limit by a subclass of DAG models ${\cal Q}_n$ with just three layers, more precisely $|{\cal P}_n| \leq (1 + \frac{1}{o(n)})|{\cal Q}_n|$. \citep{dhar:1978} investigated posets parameterized by the number $d \in (0,1)$ comparable pairs, and showed that posets can model lattice gas with energy proportional to the number of comparable pairs in the poset. Remarkably, phase transition effects appear as $d$ is varied. If we design a causal study that generates data from a uniformly randomly sampled DAG  ${\cal D} \sim P({\cal P}_n)$  over the space of {\em all} DAG models,  in the range $d \in (0, \frac{1}{8})$, as $n \rightarrow \infty$, Dhar's results imply that almost surely ${\cal D} $ is a two-layer DAG. 

Semantic causal entropy, in contrast, attempts to quantify the reverse evolutionary process of decay of the proportion of edges $d n^2$ due to causal interventions that eliminate edges. A variety of information-theoretic metrics have been proposed to quantify causal influence \citep{DBLP:conf/isit/MasseyM05,DBLP:conf/allerton/Raginsky11,nihat,wieczorek}. We generalize the edge-centric model proposed by \citep{janzing}, and define causal influence of removing a set of edges $S$ as the $\phi$ divergence $D_{\phi}(P \parallel P_S)$ between the  distribution $P$ represented by the original DAG with $P_S$ denoting the distribution represented by the DAG with edges $S$ removed. If $\phi(t) = t \log t$, we recover the model proposed by \citep{janzing}. For $t = 2 (1 - \sqrt{t})$, we get the intervention model proposed by \citep{pmlr-v65-daskalakis17a}, and for $t = \frac{1}{2}|t-1|$, we get the model proposed by \citep{acharya}. These choices lead to different decomposability properties for quantifying causal influence.  We generalize recent bipartite designs for studying causal inference on large datasets, where interventions are carried out on one set of variables (e.g., power plants, items in an online store), but outcomes are measured on a disjoint set of variables (residents near power plants, or shoppers) \citep{DBLP:journals/geoinformatica/LiFZMTZ20,DBLP:conf/nips/Pouget-AbadieAS19,DBLP:conf/kdd/Schlosser018,DBLP:conf/ijcai/SchlosserW0U18,DBLP:journals/entropy/RodderDKLR19,DBLP:conf/sigecom/CharlesCDJS10,zigler2018bipartite}. We propose a class of $k$-partite architectures, where the variables are partitioned into $[n] = X_1 \sqcup \ldots X_k$ antichains, and describe an optimization framework for finding the $k$-level DAG architecture for any value of $d \in (0, \frac{1}{2})$ that maximizes causal entropy, which reveals a sequence of phase transitions  occur for disjoint intervals of $d$, with deeper DAG architectures emerging as $d \rightarrow \frac{1}{2}$.  We give a provably quantitative estimate of the number of samples needed to measure the average causal influence from a set of edges for $k$-partite designs.

\section{Semantic Entropy using Average Causal Influence} 

\begin{figure}[t]
\begin{minipage}{0.5\textwidth}
\vspace{0pt}
\centering
\includegraphics[scale=0.5]{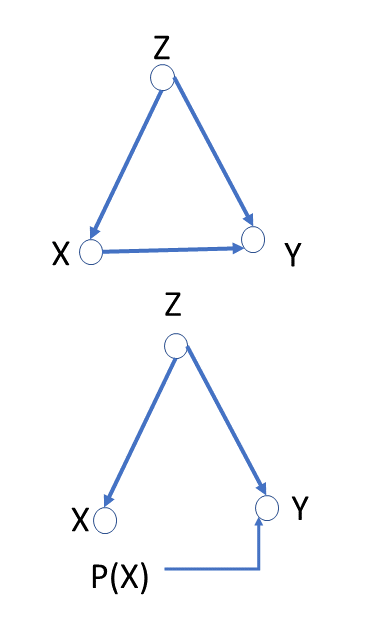}
\end{minipage}
\begin{minipage}{0.5\textwidth}
\vspace{0pt}
\centering
\begin{itemize}
    \item Causal influence ${\cal C}^\phi_{S} = D_{\phi}(P \parallel P_S)$,  for some set of edges $S$, is defined as the $\phi$-divergence between the original pre-intervention distribution $P$ with the post-intervention distribution $P_S$.
    \item For the DAG on the left, the pre-intervention distribution $P(x,y,z) = P(z) P(x | z) P(y | x,z)$, and the post-intervention distribution $P_{X \rightarrow Y}(x,y,z) = P(z) P(x | z) \sum_{x'} P(y | x',z) P(x')$. 
    \item If $\phi(t) = t \log t$, then  $C^{KL}_{X \rightarrow Y} = \sum_{x,y,z} P(x,y,z) \log \frac{P(y|x,z)}{\sum_{x'} P(y | z,x') P(x')}$, 
    \item  If both edges were included in the intervention set $T = \{(Z, X), (Z, Y) \}$, then the post-intervention distribution is simply equal to $P_T(x,y,z) = P(x) P(y) P(z)$. 
    
    \item If $X$ is simply a copy of $Z$, and $Y$ computes the XOR function of $X$ and $Z$, then if $P(Z = 0) = a, P(Z = 1) = 1 - a$, then then ${\cal C}^{KL}_S(P) = H(a)$, the entropy of $a$.
\end{itemize}
\end{minipage}
\caption{We define causal influence in terms of the $\phi$-divergence between the original distribution and the intervened distribution. We are particularly interested in choices of $\phi$, for which causal influences are decomposable. For $\phi = t \log t$, we obtain the $D_{KL}(P \parallel P_S)$ causal influence model proposed by \citep{janzing}. For $\phi(t) = \frac{1}{2}(\sqrt{t} - 1)^2$, we get the squared Hellinger distance model $D_{H^2}(P \parallel P_S)$ studied by \citep{pmlr-v65-daskalakis17a}. Finally, for $\phi(t) = \frac{1}{2}|t-1|$, we get the total variation distance model $D_{TV}(P \parallel P_S)$ studied by \citep{acharya}.  To get some intuition for the edge-centric causal influence model, in the above DAG, if $Z$ is age, $X$ is a covid vaccine, and $Y$ is getting infected, then the causal influence  ${\cal C}_{X \rightarrow Y}$ associated with removing edge $X \rightarrow Y$ measures the difference between the infection rate where vaccination rates are allowed to  depend on self-selection across age groups -- for example, older people may choose to vaccinate more than younger people -- versus the infection rate arising from a randomized vaccination strategy independent of age.} 
\label{dag1}
\end{figure}

Generally, causal discovery \citep{pearl:causalitybook,spirtes:book} is often modeled as inferring a DAG structure ${\cal G}_n = (V_n, E_n)$ \citep{pearl:bnets-book}, where $V_n = [n] = \{1, \ldots, n\}$, based on observations and interventions. Variables will be denoted in upper case, e.g. $X_i$, whereas their values will be denoted in lower case, such as $x_i$.  The edge set $E_n$ is a set $\{(x,y) \in V_n \times V_n \ |  \ (x, y) \in E_n \}$, where $x, y \in [n]$, of comparable pairs. Our focus is understanding the asymptotic case where a DAG is randomly sampled from the space of all DAGs on $n$ variables, as $n \rightarrow \infty$. We use an {\em edge-centric} intervention model (see Figure~\ref{dag1}) \citep{janzing}, which differs from the more common {\em node-centric} models used in much previous work \citep{pearl:causalitybook,DBLP:conf/uai/Eberhardt08,DBLP:journals/jmlr/HauserB12,DBLP:conf/nips/KocaogluSB17,MAOCHENG198415,prasad:aaai21,pmlr-v65-daskalakis17a,acharya}.
\begin{definition}
 Let $\Phi^*$ be the set of all convex functions $\phi(t), t \geq 0$, such that at $t = 1, \phi(1) = 0$, and at $t=0, 0 \phi(\frac{0}{0}) = 0$. The $\phi$-divergence\citep{csiszar} $D_{\phi}(P \parallel Q)$ for two discrete distributions $P$ and $Q$ is
\begin{equation}
D_{\phi}(P \parallel Q) = \sum_i q_i \phi \left(\frac{p_i}{q_i} \right) 
\end{equation}
and for the continuous case (where the $\phi$-divergence is independent of the dominating measure $\mu$):
\begin{equation}
D_{\phi}(P \parallel Q) = \int_{q > 0} \phi \left(\frac{p(x)}{q(x)} \right) \ q(x) \ d \mu(x)
\end{equation}
\end{definition}

%DAGs are constrained so that $(x, y) \in E_n$ implies $(y,x) \notin E_n$, and there are no directed cycles. A directed path $x_i, x_{i+1}, \ldots, x_j$ between variables $x_i, x_j \in [n]$ is such that each pair of nodes $(x_k, x_{k+1}) \in E$ on this path. We can define a partial ordering ${\cal P}_n$ on the vertices $[n]$ of $G_n$ from the transitive closure of the original graph ${\cal G}_n$. Bayesian networks \citep{pearl:bnets-book} and structural causal models \citep{pearl:causalitybook} are defined by probability distributions $P_{{\cal G}_n}$ on a DAG $G_n$, such that the conditional independences defined on the graph are exactly those satisfied by the probability distribution $P_{{\cal G}_n}$ \citep{lauritzen:text}. We will often denote the distribution simply as $P$, if the associated DAG is clear from the context. Given a DAG $G_n$, a set of nodes $X$ is conditionally independent of a set $Y$, given a third set $Z$, where $X, Y, Z \subset [n]$, denoted $(X \Perp Y | Z)_{{\cal G}_n}$, if each directed path from a node $x \in X$ to a node $y \in Y$ is such that for any collider node $i \rightarrow j \leftarrow k$ in the path, neither $j$ nor any of its descendants $\in Z$, and there exists some non-collider node $i \rightarrow j \rightarrow k$ in the path such that $j \in Z$. In the distribution $P$, the equivalent conditional independence property is defined as $(X \Perp Y | Z)_P$ if $P(X | Y, Z) = P(X | Z)$. A detailed analysis of conditional independence properties is given in \citep{lauritzen:text,pearl:bnets-book}. 

 \begin{definition}
 The {\bf causal influence} ${\cal C}^\phi_S$ of a set of edges $S \subset E_n$ in a DAG $G_n = ([n], E_n)$ is defined as the $\phi$-divergence $D_{\phi}(P \parallel P_S)$, for some convex function $\phi$, where $\phi(1) = 0$, and  where $P$ is the original pre-intervention distribution and $P_S$ is the post-intervention distribution, defined as follows: 
 \begin{equation}
     P(x_1, \ldots, x_n) = \prod_{i=1}^n P(x_i | \mbox{pa}_i)
 \end{equation}
 where $\mbox{Pa}_i$ is the set of parent variables of $X_i$ (and $\mbox{pa}_i$ are their specific instantiations). Given a set of edges $S$ whose causal influence is to be quantified, the post intervention distribution $P_S$ is defined as:
 \begin{equation}
 P_S(x_1, \ldots, x_n) = \prod_{i=1}^n P_S(x_i | \mbox{pa}^{\bar{S}}_i)  = \prod_{i=1}^n \sum_{\mbox{pa}^S_i} P(x_j | \mbox{pa}^{\bar{S}}, \mbox{pa}^S_i) P_{\Pi}(\mbox{pa}^S_j)
 \end{equation}
 where $S = \{(i,j) \in E_n \}$ is the set of edges intervened on, $\bar{S} = \{(i,j) \notin S \}$ are the non-intervened edges, $\mbox{Pa}^S_i$ is the set of parents of variable $X_i$ whose edges are included in $S$, $\mbox{Pa}^{\bar{S}}_i$ are the parents of variable $X_i$ whose edges are not included in $S$, and $P_{\Pi}(\mbox{pa}^S_i)$ is the product of marginal distributions of all variables in $\mbox{Pa}^S_i$. 
 \end{definition}
%
%A detailed discussion of the properties of causal influence is given in \citep{janzing}, which briefly we can summarize as follows. (i) Causal influences satisfy a causal Markov condition: if ${\cal C}_S(P) = 0$, then the intervened distribution $P_S$ satisfies the Markov condition with respect to the DAG $G_S$ that has the edges corresponding to $S$ removed. (ii) For a DAG with a single edge $X \rightarrow Y$, ${\cal C}_{X \rightarrow Y} = I(X; Y)$, the mutual information of $X$ and $Y$. (iii) Causal influences are decomposable with respect to the edges that share a common target variable. (iv) Causal influence is hereditary, namely if $C_{\cal S}(P) = 0$, then this condition is satisfied for any subset of edges $T \subset S$. 
%
\begin{definition}
The causal influence $C_{X \rightarrow Y}(P)$ is {\bf localizable}  if the strength of $X \rightarrow Y$ only depends on knowing $P(Y | \mbox{PA}_Y)$ and $P(\mbox{PA}_Y)$.
\end{definition}
\begin{definition}
Given a set of edges $S$, $\mbox{trg}(S) = \{X_i | (X_j, X_i) \in S \}$ denotes the {\bf target nodes of the intervention}.
\end{definition}
\citep{janzing} defined causal influence and localizability specifically for $D_{KL}(P \parallel P_S)$. Below we show that this notion can be significantly generalized to several other types of $\phi$ divergences, which 
lead to different decomposability properties, as summarized by the following result. 
\begin{theorem}
 The causal influence ${\cal C}^\phi_S = D_{\phi}(P \parallel P_S)$ of a set of edges $S \subset E_n$ in a DAG $G_n = ([n], E_n)$ has the following decomposability properties, depending on choice of $\phi$:
\begin{itemize}
    \item $\phi = t \log t$: Using the chain rule for KL-divergences, it can be shown that $C^{KL}_S(P)  = \sum_i  D_{KL}(P \parallel P_{S_i}) = \sum_{X_i \in \mbox{trg}(S)} D_{KL}(P(X_i | PA_i) \parallel P_S(X_i \parallel \mbox{PA}_i)$. This result was shown in \citep{janzing}. 
    \item $\phi = \frac{1}{2}(\sqrt{t} - 1)^2$ or $t = \frac{1}{2} |t-1|$: In these cases, corresponding to squared Hellinger distance and total variation distance,  the results of \citep{pmlr-v65-daskalakis17a} and \citep{acharya} imply that the resulting causal influences are {\em subadditive}
    \begin{equation}
        {\cal C}^\phi_S = D_{\phi}(P \parallel P_S) \leq \sum_{X_i  \in \mbox{trg}(S)} D_{\phi}(P_{X_i \cup \mbox{Pa}_i} \parallel P_{X_i \cup \mbox{Pa}^{\bar{S}}_i})
    \end{equation}
\end{itemize}
\end{theorem}
\citep{DBLP:conf/aistats/DingDF21} give a detailed analysis of subadditivity properties of various other $\phi$-divergences, including $\chi^2$, Wasserstein distance, Jensen-Shannon divergences and several others. 
As we show later, a key strength of squared Hellinger distance and total variation distance over KL divergence is that the former metrics provide sample-efficient testing methods \citep{acharya,pmlr-v65-daskalakis17a}. 
A well-known result relates the above three $\phi$ divergences, showing they are closely related. 
\begin{lemma}
For any two distributions $P$ and $Q$, the following well-known inequalities hold \citep{DBLP:journals/corr/SasonV15}:
\begin{equation}
D_{TV}(P \parallel Q) \leq \sqrt{2} \ D_{H^2}(P \parallel Q) \leq \sqrt{D_{KL}(P \parallel Q)}
\end{equation}
\end{lemma}
Similar to the standard notion of average treatment effect in the Neyman-Rubin potential outcomes framework \citep{rubin-book}, we can define the average causal influence of a set of edges $S$: 
\begin{definition}
{\bf Average causal influence} $ACI^\phi(S)$ of a set of edges $S \subset E_n$ in a DAG $G_n = ([n], E_n)$ is defined as the average $\phi$-divergence  $\frac{1}{|S|} D_{\phi}(P \parallel P_S)$, where $P$ is the original pre-intervention distribution and $P_S$ is the post-intervention distribution. 
\end{definition}
%

%One of the most widely used algorithms for estimating ATE is the classic Horvitz-Thompson (HT) method, developed in the mid 1950s \citep{DBLP:reference/stat/Maiti11}. The HT estimator has been the subject of numerous studies over many decades characterizing its bias and variance. Recently, an interesting Gram-Schmidt random walk algorithm \citep{gs-random-walk} has been proposed based on the HT estimator, which builds on work in theoretical computer science on discrepancy theory \citep{DBLP:conf/stoc/BansalJ0S20}. 

\section{Causal Structural Entropy and Phase Transitions} 

Causal edge interventions remove edges, and reduce the number of comparable relations. We now look at the growth of edges as the proportion of comparable relations $d$ increases from $0$ to $\frac{1}{2}$. Structural entropy of a causal model captures its description complexity, defined as $c(d) \equiv \lim_{n \rightarrow \infty} \frac{\log_2 | {\cal P}^d_n |}{n^2}$, the limit of the logarithmic growth rate, measured in bits, of a directed acyclic graph (DAG) (or equivalently, the induced partially ordered set) on $n$ variables ${\cal P}^d_n$, parameterized by the relative proportion $d \in (0,1)$ of the total $d n^2$ edges in the DAG (or relations in the poset).  It is well known that the set of possible DAG structures on $n$ variables grows superexponentially \citep{robinson}. Surprisingly, it turns out that longstanding results in extremal combinatorics on the structure of partially ordered sets \citep{dhar:1978,dhar:1980,kleitman:1975,promel:phase} gives insights into the structure of DAG models in high-dimensional spaces. 
%\citep{kleitman:1975} pioneered a novel approach to extremal combinatorics in their study of the global structure of partially ordered sets ${\cal P}_n$, when they showed it could be decomposed into a disjoint collection of about $15$ different subsets, such that asymptotically, as $n \rightarrow \infty$, surprisingly a few simple subsets ${\cal Q}_n$ dominated the total space. 

\begin{definition}
Let ${\cal P}_n$ denote the family of {\bf labeled posets} on the point set $[n]$, where a particular poset $P \in {\cal P}_n$ is defined as a set of {\bf comparable pairs} $(x,y) \in P$ iff $x \leq y$, and $x < y$ if $x \leq y$ and $x \neq y$. We say $x$ is {\bf covered} by $y$ if $x < y$ and there is no other element $z$ such that $x < z$ and $z < y$. The {\bf cover graph} associated with a poset $P$ is the directed graph $G_P = ([n], E_P)$ whose edges $E_P$ are defined by the cover relationship of the poset. 
\end{definition}

\begin{definition}
The {\bf Hasse diagram} ${\cal H}_P$ of a poset $P$ is a DAG with vertices $[n]$, and a single directed edge from $a \rightarrow b$ if and only if $a$ covers $b$. Distinct posets in ${\cal P}_n$ define distinct graphs. A Hasse diagram is in fact a DAG, as it cannot have any directed cycles, which would violate transitivity. In particular, Hasse diagrams are defined to contain no triangles either, that is $a \rightarrow b, b \rightarrow c, a \rightarrow c$.  A node $a$ is {\bf adjacent} to $b$ in a Hasse diagram if $a$ covers $b$ or $b$ covers $a$. 
\end{definition}

\begin{definition}
The {\bf levels} of a Hasse diagram ${\cal H}_P$ of a poset $P$ is defined as follows. Level 1 consists of all minimal vertices of ${\cal H}_P$, that is vertices that are not covering any other vertex. For each $j > 1$, level $j$ is the set of minimal vertices obtained by deleting all vertices in levels $1, \ldots, j-1$. If $a$ is in level $i$, and $b$ is in level $j$, if $i < j$, then either $a < b$ or $a$ and $b$ are incomparable. Two vertices at the same level are necessarily incomparable, consequently vertices at a level form an antichain. 
\end{definition}

\begin{definition}
Given a poset $P$ on $[n]$, a {\bf chain} ${\cal C} \subset [n]$ is defined to be a totally ordered subset of $P$. The {\bf height} of a poset is defined as the maximum cardinality of a chain. An {\bf antichain}  ${\cal A} \subset [n]$ of a poset $P$ is a subset of elements in which no pair of elements are ordered. The set of all elements $[n]$ can be partitioned into disjoint subsets of chains or antichains, whose sizes are related by Mirsky's theorem \citep{mirksy:dilworth}, which simply states that the number of antichains is lower bounded by the number of chains, as no two elements of a chain can ever be in an antichain. 
\end{definition}

%To connect these ideas to recent work on causal discovery of DAGs, we note that Algorithm 2 proposed by \citep{DBLP:conf/nips/KocaogluSB17} is based on using the antichain sets of an induced partial ordering as intervention sets to discover the underlying DAG structure. Their algorithm depends on the height of the partial ordering, and a nice theoretical connection exists between height and the size of antichains. Our theoretical framework has implications for DAG recovery algorithms such as these, based on theoretical estimates for the optimal $k$-layer poset architecture in the asymptotic regime as $n \rightarrow \infty$. 

\begin{theorem}
{\bf Mirsky's theorem} \citep{mirksy:dilworth}: The {\bf height} of a partial ordering $P$ is defined to be the maximum cardinality of a chain, a totally ordered subset of the given partial order. For every partial ordering $P$, its height also equals the minimum number of antichains. 
\end{theorem}

The challenge in combinatorial enumeration is obtaining good upper bounds on the size of ${\cal P}_n$, but lower bounds are of course much easier. 

\begin{theorem}
The number of posets ${\cal P}_n$ on a finite set $[n]$ is $|{\cal P}_n | \geq 2^{\frac{n^2}{4}}$. 
\end{theorem}

{\bf Proof:} Fix two antichains $X$ and $Y$, each on $\frac{n}{2}$ points, and for each of the $\frac{n^2}{4}$ comparable pairs $(x, y) \in X \times Y$, decide if $x \leq y$ or $y \leq x$. $\qed$

Upper bounding the number of posets ${\cal P}_n$ is much harder. To develop some intuition, let us consider a canonical representation of posets that will be used in the remainder of the paper. 
\begin{definition}
\label{kposet}
Define the class of {\bf $k$-partitioned models (DAGs or posets)} as ${\cal Q}^k_n$ where the elements $[n] = V_1 \sqcup V_2 \ldots V_k$ form a disjoint class of $k$ partitions, satisfying the following conditions: 
\begin{enumerate}
    \item If $x \in V_i$ and $y \in V_j$, with $x < y$, then $i < j$. 
    \item If $x \in V_i$ and $y \in V_j$ and $i < j-1$, then $x < y$. 
\end{enumerate}
\end{definition}
Condition 1 above ensures the antichain elements are not comparable. Condition 2 restricts the poset so that each element at level $L_i$ is comparable to every element at level $L_{i+2}$, where $1 \leq i \leq k-2$. This second restriction is imposed to make the enumeration simpler, but as it will turn out, these restricted models in fact completely characterize the space of {\em all} posets on $n$ variables, as $n \rightarrow \infty$.
\begin{theorem}
The number of $k$-partite causal models is $|{\cal Q}^k_n| \sim 2^{\sum_{i=1}^{k-1} | V_i | |V_{i+1} |}$
\end{theorem}

{\bf Proof:} Given the conditions imposed by Definition~\ref{kposet}, the only freedom in generating a poset is choosing the relations between the elements in $V_i$ and $V_{i+1}$, for $1 \leq i \leq k-1$. Note that we are ignoring the number of ways of distributing the elements in each antichain, as it only adds factors of size $o(n)$ to the exponent. $\qed$

A useful property that links combinatorics and entropy is given by the following lemma. 

\begin{lemma}
\label{entropy-comb}
For any fixed $0 < x < 1, {n \choose n x}  = 2^{H(x) n + o(n)}$. 
\end{lemma}

\begin{theorem}
\citep{kleitman:1975} proved the following classic upper bound: 
\begin{equation}
\label{pbound}
  \log_2 |{\cal P}_n | = \frac{n^2}{4} + \frac{3n}{2} + \mathcal{O}(\log_2 (n))
\end{equation}
\end{theorem}

{\bf Sketch of Proof:} \citep{kleitman:1975}  classified the set of all DAG models (Hasse diagrams of posets) in ${\cal P}_n$ into $15$ disjoint classes, and show that the class of all models, in the limit as $n \rightarrow \infty$, was dominated by one particular subclass of DAGs ${\cal Q}_n$ with exactly three levels defined below. $\qed$

\begin{definition}
\label{3poset}
The class of {\bf three layer DAG models} ${\cal Q}_{m+1}$ on $[m+1]$ elements is defined as follows. 
\begin{enumerate}
    \item The set of variables $V$ in ${\cal Q}_{m+1}$ is decomposed as $V = V_1 \sqcup V_2 \sqcup V_3$, where $V_1$, $V_2$, and $V_3$ are the three antichains of the poset. 
    
    \item For levels $i = 1,3$, $\frac{m+1}{4} - \sqrt{m} \log m < |V_i| < \frac{m+1}{4} + \sqrt{m} \log m$
    
     \item For level $2$, $\frac{m+1}{2} - \log m < |V_2| < \frac{m+1}{2} + \log m$
    
\end{enumerate}
\end{definition}

\begin{theorem}
The set of all posets ${\cal P}_n$ is such that $|{\cal P}_n | = (1 + \mathcal{O}(\frac{1}{n}) |{\cal Q}_n |$. Informally, {\bf all causal models in the limit are essentially three layer DAGs!}. 
\end{theorem}

%\subsection{Phase Transitions} 

\begin{figure}[t]
\centering
\begin{minipage}{1\textwidth}
\centering
\includegraphics[scale=0.35]{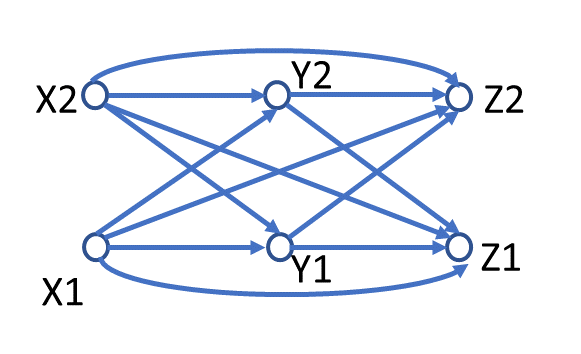}
\includegraphics[scale=0.1]{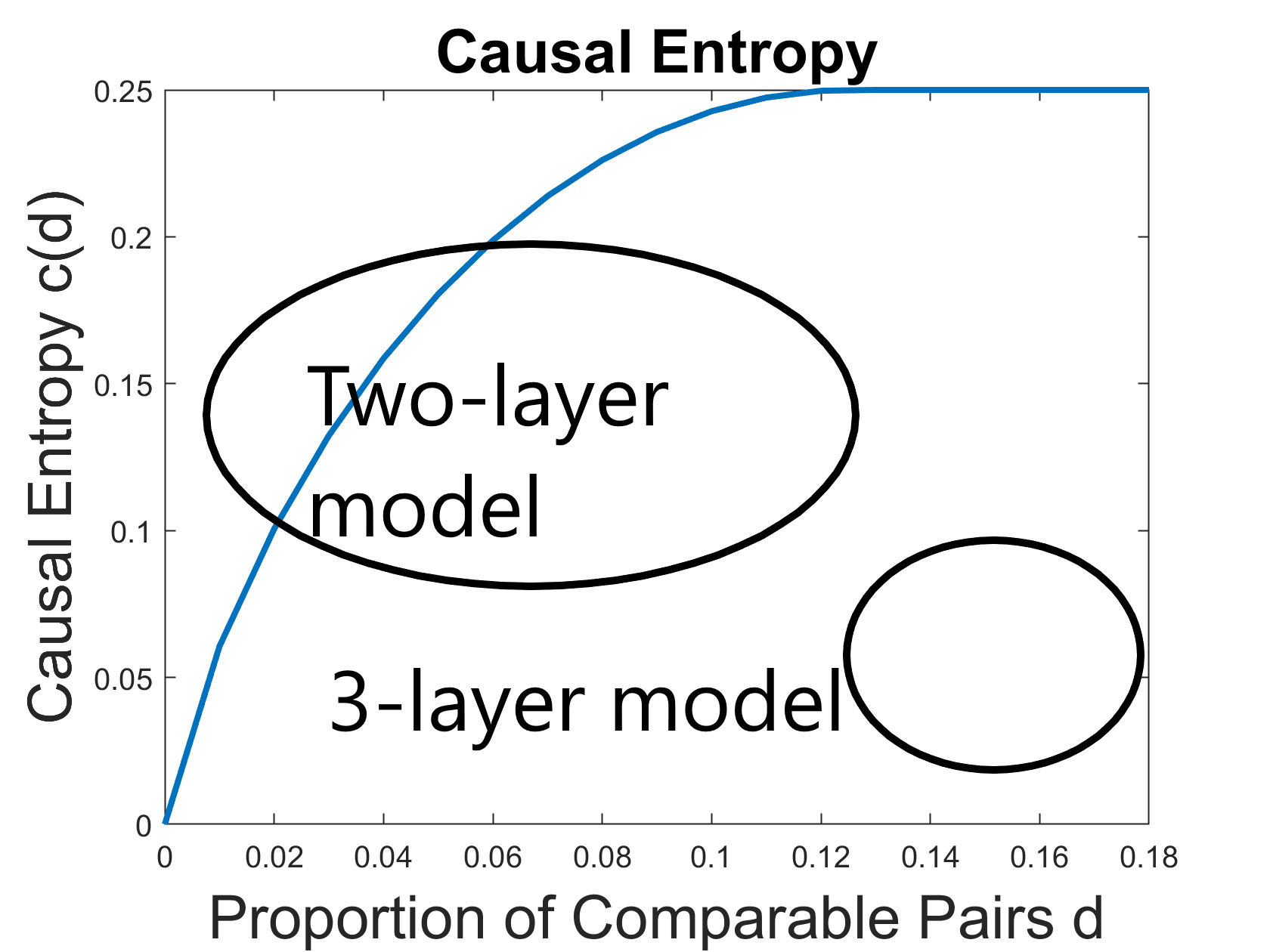}
\end{minipage}
\caption{(Left) \citep{kleitman:1975} showed that all posets on $n$ variables, as $n \rightarrow \infty$, are asymptotically dominated by the subclass of 3-layer posets, comprised of 3 antichains of sizes $|V_1| = |V_3| = O(\frac{n}{4}), |V_2| = O(\frac{n}{2})$ (see Definition~\ref{3poset}). (Right) \citep{dhar:1978} investigated poset models parameterized by $d n^2$ edges, and discovered phase transitions in their evolution over $d \in (0, \frac{1}{2})$. In the range $0 < d \leq \frac{1}{8}$, two layer posets dominate, whereas in the range $\frac{1}{8} \leq d \leq \frac{3}{16}$, three-layer posets are dominant. The plot shows the structural causal entropy function $c(d)$, over $d \in (0, \frac{3}{16}).$} 
\label{diagram2}
\end{figure}

We now define {\em structural causal entropy}, based on the parameterization of DAG and poset models using the percentage $d \in (0,1)$ of comparable elements,  Remarkably, phase transition effects emerge, as $d$, the proportion of comparable items, is varied (see Figure~\ref{diagram2}), a result first noted by \citep{dhar:1978} connecting statistical physics and the structure of posets.  

\begin{definition}
Define the space of all causal structures (DAGs or posets) ${\cal P}^d_n$, where $d \in (0,1)$, such that in any model $P \in {\cal P}^d_n$, only $d n^2$ pairs are comparable. The {\bf structural causal entropy} of a model $P \in {\cal P}^d_n$, with $d n^2$ comparable relations, where $d \in (0,1)$  is defined as \citep{dhar:1978,promel:2001}:
\begin{equation}
    c(d) = \lim_{n \rightarrow \infty} \frac{\log_2 |{\cal P}^d_n |}{n^2}
\end{equation}
provided the limit exists. 
\end{definition}

The reason for using the term entropy in this context is that the posets can represent the states of a certain model of a lattice gas with energy proportional to the number of comparable pairs. Given the upper bound given by Equation~\ref{pbound}, the following theorem is straightforward to prove. 
\begin{theorem}
The structural causal entropy of a poset $P \in {\cal P}^d_n$ on $[n]$ is upper bounded by 
\[ c(d) \leq \frac{1}{4} \] 
\end{theorem}

\begin{theorem}\citep{dhar:1980}
(i) The function $\frac{c(d)}{d}$ is a monotonic nonincreasing function of $d$. (ii) The function $\frac{c(d)}{1-d}$ is a monotonic nondecreasing function of $d$. 
\end{theorem}

Remarkably, as $d \in (0,1)$ is varied, \citep{dhar:1978} noticed interesting phase transition effects occur, which we will explore below. In particular, \citep{dhar:1978} was able to show that in the range $\frac{1}{8} \leq d \leq \frac{3}{16}$, the structural causal entropy $c(d) = \frac{1}{4}$. It is worth understanding this result in more depth. 

\begin{theorem}
\label{dhar-thm}
For $d$ in the range $\frac{1}{8} \leq d \leq \frac{3}{16}$, the causal entropy $c(d) = \frac{1}{4}$.
\end{theorem}

{\bf Proof:} Let us define $x = \frac{1}{4} - \sqrt{\frac{3}{16} - d}$. Then, as $\frac{1}{8} \leq d \leq \frac{3}{16}$, it must be the case that $0 \leq x \leq \frac{1}{4}$. To show the theorem, we have to can construct a DAG of $3$ layers, with $[n] = V_1 \sqcup V_2 \sqcup X_3$, where $|V_1| = (\frac{1}{2} - x) n$ elements, $|V_2| = \frac{1}{2} n$ elements, and $|V_3| = x n$ elements. For the edges across the layers, insert $\frac{1}{2} |V_i| |V_{i+1}|$ edges between layers $L_i$ and $L_{i+1}$, for $i=1,2$, and all possible edges between $L_1$ and $L_3$. This can be shown to give a total of $d n^2$ relations, hence this DAG will achieve the entropy $c(d) = \frac{1}{4}$ as $n \rightarrow \infty$. $\qed$

\begin{theorem}
The structural entropy of a model $P \in {\cal P}^d_n$ on $[n]$ for any value of $d \in (0,1)$ in the following ranges is given by: 
\begin{itemize}
    \item  $0 < d \leq \frac{1}{8}$: In this range for $d$, $c(d) = \frac{1}{4} {\cal H}(4 d)$, where ${\cal H}(x) = -x \log_2 x - (1 - x) \log_2 (1 - x)$, the entropy function. 
    
    \item $\frac{1}{8} \leq d \leq \frac{3}{16}$: In this range for $d$, $c(d) = \frac{1}{4}$, the entropy is constant (as shown in Theorem~\ref{dhar-thm}). 
\end{itemize}
\end{theorem}

Figure~\ref{diagram2} plots the value of the causal entropy $c(d)$ over the range $0 \leq d \leq \frac{3}{16}$. A phase transition occurs between the ranges $0 < d \leq \frac{1}{8}$, when a two-layer poset dominates the space asymptotically, and the range $\frac{1}{8} \leq d \frac{3}{16}$, when as shown by Theorem~\ref{dhar-thm}, a three-layer poset architecture dominates the space ${\cal P}^d_n$.  

\section{Algorithms for Designing and Testing Multi-partite Causal Designs} 

We now use the above information-theoretic framework to generalize recent studies of causal inference on large datasets that use  {\em bipartite experiments} \citep{DBLP:journals/geoinformatica/LiFZMTZ20,DBLP:conf/nips/Pouget-AbadieAS19,DBLP:conf/kdd/Schlosser018,DBLP:conf/ijcai/SchlosserW0U18,DBLP:journals/entropy/RodderDKLR19,DBLP:conf/sigecom/CharlesCDJS10}. Bipartite experiments are characterized by two types of units, {\em interventional units} ${\cal T} = \{p_1, \ldots, p_k \}$, -- for example, power plants, teachers, items in a marketplace, and neighborhoods -- and {\em outcome units} ${\cal O} = \{o_1, \ldots, o_m \}$, such as health outcomes of people living near power plants, students in a classroom, prospective buyers in an online marketplace, and residents of a neighborhood. 

\begin{definition}
{\bf Average causal influence} $ACI^\phi(S)$ of a set of edges $S \subset E_n$ in a bipartite DAG $G_n = (V_1 \sqcup V_2, E_n)$, where the set of all nodes is partitioned into two disjoint subsets (antichains) $V_1$ and $V_2$, where $V_1$ are the {\em interventional} units and $V_2$ are the {\em outcome units}.
\begin{itemize}
    \item For $\phi(t) = t \log t$, we get
    \begin{equation}
ACI^{KL}(S,2) = \frac{1}{|S|} \sum_{X_i \in \mbox{trg}(S \cap V_2)} D_{KL}(P(X_i | PA_i) \parallel P_S(X_i \parallel \mbox{PA}_i)
\end{equation}
where $S_i \in \mbox{trg}(S \cap V_2)$ is the set of outcome nodes in layer 2 that are in  $S$. 

\item For $\phi(t) = \frac{1}{2} |t-1|$, or $\phi(t) = \frac{1}{2}(\sqrt{t} - 1)^2$, the resulting decomposition of average causal influence is: 
 \begin{equation}
        {\cal C}^\phi_S = D_{\phi}(P \parallel P_S) \leq \sum_{X_i \in  \mbox{trg}(S \cap V_2)}  D_{\phi}(P_{X_i \cup \mbox{Pa}_i} \parallel P_{X_i \cup \mbox{Pa}^{\bar{S}}_i})
\end{equation}
\end{itemize}

where $P$ is the original pre-intervention distribution and $P_S$ is the post-intervention distribution. 
\end{definition}

\subsection{Structural Entropy of Multi-partite designs} 

The two-layer poset architecture in bipartite design can be explained in terms of our structural entropy framework by estimating the empirical value of the number of comparable relations $d$ in these domains. For example,  \citep{harshaw2021design}, \citep{DBLP:conf/sigir/McAuleyTSH15} and others study online experimentation on a large Amazon dataset, where there are over $83$ million reviews made by $121,000$ thousand reviewers on $9.8$ million items. Out of the total set of possible comparable pairs (item reviewer combinations), which would be $9.8 x 10^6 \times 121,000$ = $1.1858 x 10^{12}$ possible comparable pairs,this dataset has only $83$ million comparable items, which translates to the value of $d = \frac{83 \times 10^6}{9.8 \times 10^6 \times 121,000} \approx \frac{1}{100,000} = 10^{-5}$. Since $d$ is close to $0$, the theory confirms that bipartite structures are appropriate. At the threshold point of $d > \frac{1}{8}$, it may be more appropriate to use a tri-partite design, and furthermore, as $d > \frac{3}{16}$, more than $3$ levels may be desirable. To generalize bipartite designs, we note any causal model in ${\cal P}^d_n$ can be well approximated by a $k$-layer model in ${\cal Q}^d_n$ \citep{promel:phase}. The proof uses Szemer\"{e}di's regularity lemma  \citep{szemeredi:1975}. 
\begin{definition}
The {\bf parameterized subclass of $k$-layer poset models} ${\cal Q}^d_n = (\lambda_1, \ldots, \lambda_k, p)$ is defined as one where the variables are partitioned into $X = V_1 \sqcup \ldots V_k$ antichains, with $p |V_i| |V_{i+1}|$ comparable pairs across layers $i, i+1$, $|V_i| = \lambda_i n$, and $\sum_{i=1}^n \lambda_i = 1$.
\end{definition}
\begin{theorem}\citep{Taraz1999Phase}
For every $\epsilon > 0$, and every $0 < d < \frac{1}{2}$, there exists constants $k_0, n_0$ such that, for every poset $P \in {\cal P}^d_n$ with $n \geq n_0$, there exists a $k$-partitionable poset $P' \in {\cal Q}^d_n$ with $k \leq k_0$ that differs from $P$ in atmost $\epsilon n^2$ relations, and in which the partition classes differ in size by at most one. 
\end{theorem}
\begin{definition}
\label{kpartite-aci}
{\bf Average causal influence} $ACI^\phi(S,k)$ of a set of edges $S \subset E_n$ in a $k$-level DAG $G_n = (V_1 \sqcup \ldots V_k, E_n)$, where the set of all nodes is partitioned into $k$ disjoint subsets (antichains) is defined as 
\begin{itemize}
    \item For $\phi(t) = t \log t$, we get
    \begin{equation}
    ACI^{KL}(S,k) = \frac{1}{|S|} \sum_{j=2}^k \sum_{X_i \in \mbox{trg}(S \cap V_j)} D_{KL}(P(X_i | PA_i) \parallel P_S(X_i \parallel \mbox{PA}_i)
\end{equation}
where $X_i \in \mbox{trg}(S \cap X_i)$ is the set of outcome nodes in layer 2 that are in the target set of $S$. 

\item For $\phi(t) = \frac{1}{2} |t-1|$, or $\phi(t) = \frac{1}{2}(\sqrt{t} - 1)^2$, the resulting decomposition of average causal influence is: 
 \begin{equation}
        ACI^{\phi}(S,k) = \frac{1}{|S|}  D_{\phi}(P \parallel P_S) \leq \frac{1}{|S|} \sum_{j=2}^k \sum_{X_i \in \mbox{trg}(S \cap V_j)} D_{\phi}(P_{X_i \cup \mbox{Pa}_i} \parallel P_{X_i \cup \mbox{Pa}^{\bar{S}}_i})
\end{equation}
\end{itemize}

where $P$ is the original pre-intervention distribution and $P_S$ is the post-intervention distribution. 
\end{definition}
Algorithm 1 specifies a procedure to design a $k$-level DAG architecture that maximizes structural causal entropy. This procedure is based on the theory of evolution of posets \citep{promel:phase,Taraz1999Phase}.  Note that the number of $k$-partite designs $|Q^n_d|$ is bounded by $\sum_{i=1}^{k-1} {\lambda_i \lambda_{i+1} n^2 \choose p \lambda_i \lambda_{i+1} n^2}$, ignoring the ways in which variables can be assigned to layers, which adds only a $2^{\mathcal{O}(n \log n)}$ that is negligible under the limit $\lim_{n \rightarrow \infty} \frac{\log_2 |Q^n_d|}{n^2}$.
\begin{algorithm}[h]
\caption{Given a dataset ${\cal D}$ on $[n]$ features, where $n \rightarrow \infty$ (meaning $n$ is very large), determine the optimal $k$-level poset architecture  $P \in {\cal P}^d_n$ on $[n]$.}
\SetAlgoLined
\KwIn{A dataset ${\cal D}$ on $[n]$ features, where $n \rightarrow \infty$ (meaning $n$ is very large).}
\KwOut{A poset $P \in {\cal P}^d_n$ on $[n]$, with the induced partial ordering $\leq$. }
\Begin{
    Estimate edge density $d$ by sampling relations from the dataset ${\cal D}$.  \\ 
    Solve the optimization problem in Theorem~\ref{klayer-opt} to determine the asymptotically optimal $k$-layer poset architecture $P \in {\cal Q}^d_n$, where $P = (\lambda_1, \ldots, \lambda_k, p)$ specifies the elements at each layer, and the relations among them.  \\ 
    Return the poset $P$. 
}
\end{algorithm}

\begin{theorem} \citep{Taraz1999Phase}
\label{klayer-opt}
The {\bf maximum entropy $k$-layer poset architecture} ${\cal Q} = (\lambda_1, \ldots, \lambda_k, p)$ can be found by solving the following optimization problem: 
\begin{itemize}
    \item Choose $k, \lambda_1, \ldots, \lambda_k$ and $p$ such as to maximize $H(p) \sum_{i=1}^{k-1} \lambda_i \lambda_{i+1}$, 
    \item subject to $\frac{1}{2} - \frac{1}{2} \sum_{i=1}^k \lambda_i^2 - (1 - p) \sum_{i=1}^{k-1} \lambda_i \lambda_{i+1} = d$ 
    \item where $\sum_{i=1}^k \lambda_i  = 1, \ \ 0 < \lambda_i < 1, \ \  \frac{1}{2} \leq p \leq 1$. 
\end{itemize}
\end{theorem}

\subsection{Estimating Causal Influence for Multi-partite designs} 

\begin{lemma}
{\bf Hellinger Test:}\citep{DBLP:conf/nips/AcharyaDK15,acharya,pmlr-v65-daskalakis17a}. From $\Tilde{\mathcal{O}}(D^{\frac{3}{4}}/\epsilon^2)$ samples of distributions $P$ and $Q$ over the same finite set of size $D$,  we can distinguish between $P=Q$ vs. $D_{H^2}(P \parallel Q) \geq \epsilon^2$ with error probability at most $1/3$. The error probability can be made smaller than $\delta$ with an additional factor of $\mathcal{O}(\log \frac{1}{\delta})$ in sample complexity. 
\end{lemma}
\begin{algorithm}[h]
\caption{Given a $k$-layer DAG  ${\cal Q} = (\lambda_1, \ldots, \lambda_k, p)$ with a distribution $P$, and a set of edges $S$, test whether the causal influence $C^\phi(S) \geq \epsilon$.}
\SetAlgoLined
\KwIn{$k$-layer DAG model ${\cal Q} = (\lambda_1, \ldots, \lambda_k, p)$ with distribution $P$, and a set of edges $S$, and threshold $\epsilon$.}
\KwOut{Test if causal influence $C^\phi(S) \geq \epsilon$.}
\Begin{
     For each of $\omega = |V_k \cap S|$ outcome nodes, run a squared Hellinger test, where $d$ is the maximum number of parents of any outcome node at layer $k$,  $|\Sigma|$ is the size of the discrete alphabet. \\
     \For{variable $X_i \in S \cap V_k$ at layer $k$}{
	 Run the Hellinger test to distinguish between 
	 \begin{itemize}
    \item $H_0$: $P_{X_i \cup \mbox{Pa}_i} = P_{X_i \cup \mbox{Pa}^{\bar{S}}_i}$
    \item $H_1$: $D_\phi(P_{X_i \cup \mbox{Pa}_i} \parallel P_{X_i \cup \mbox{Pa}^{\bar{S}}_i}) \geq \frac{\epsilon^2}{2 \omega}$ 
    \end{itemize}
	 }
	 
    If the null hypothesis $H_0$ succeeds in all tests, return failure, else return success.
}
\end{algorithm}
%We now present a novel framework for determining causal influence for $k$-partite designs, which unlike the previous work on bipartite experiments, gives provably guaranteed sample and time complexity bounds. The key idea is to use the subaddivity of the squared Hellinger and total variation distance distance $D_{H^2}(P \parallel P_S)$ and $D_{TV}(P \parallel P_S)$, where $P$ denotes the distribution when all intervention units are in ``control" mode, and $P_S$ is the intervention distribution that results when the edges $S$ between treatment and outcome units are intervened upon. Our estimator for  causal influence is based on recent work on hypothesis testing for Bayes nets based on the squared Hellinger distance $\citep{acharya,pmlr-v65-daskalakis17a}$. 
%
\begin{theorem}
Algorithm 2 computes the causal influence $C^\phi(S)$ of a set of edges $S$ in a $k$-partite design (see Definition~\ref{kpartite-aci}), where $\phi(t) = \frac{1}{2}(\sqrt{t} - 1)^2$ in time $\Tilde{\mathcal{O}}(\frac{|\Sigma|^{\frac{3}{4}(d+1)}\omega}{\epsilon^2})$, where $\omega = |S \cap V_k|$, by running $\omega$ squared Hellinger tests, where $d$ is the maximum number of parents of any outcome node at layer $k$, and $|\Sigma|$ is the size of the discrete alphabet.
\end{theorem}
{\bf Proof:} We exploit the subaddivity of the squared Hellinger distance. We assume a $k$-partite design where all $|V_k| = \lambda_k n$ outcome units are placed at the bottom $k$th layer, thus we only need to test $\omega = S \cap |V_k|$ of them to determine the causal influence.  Let $p_1, \ldots, p_\omega$ denote the marginals over the outcome units without intervention, and $q_1, \ldots, q_\omega$ denote the corresponding marginals under intervention on the edges in $S$. Then, if the causal influence of the edges $S$ $D_{H^2}(p,q) \geq \epsilon$, subadditivity implies that $\sum_{i=1}^\omega \frac{(p_i - q_i)^2}{q_i} \geq \frac{\epsilon^2}{2}$, which is the $\chi^2$ divergence over the corresponding vectors $p$ and $q$, and the results from \citep{DBLP:conf/nips/AcharyaDK15} can be used.  $\qed$ 

We can adapt Algorithm 2 to test for location differences (e.g., shift of the means) in the original and intervened distributions by computing the (sample) expectations, and checking the average treatment effect of the intervention. We can also extend Algorithm 2 to the continuous case, using a linear structural equation model (SEM) to model the outcome of each unit at layer $k$ given a particular exposure to treatment units. In this SEM case, we can view outcomes as a multivariate Gaussian, and we can compare the resulting multivariate Gaussian distributions using the squared Hellinger distance $D_{H^2}(P \parallel P_S)$ for which a closed form expression is well-known.

\section{Power-Divergence based Hypothesis Testing of Causal Influence}

In this paper, we defined semantic causal entropy as the $\phi$-divergence \citep{ali-silvey,csiszar} between the original distribution $P$ and the intervention distribution $P_S$ over a causal model ${\cal G} = (V, E, P)$ represented as a directed acyclic graph (DAG), where $S \subset E$ is a subset of directed edges whose causal influence is to determined, and $P$ is a probability distribution that is Markov w.r.t. to the conditional independences in the graphical model. We proposed a way to estimate causal influence given a set of samples from both the original distribution $P$ and the intervened distribution $P_S$, building on the $\chi^2$-based Bayes network hypothesis testing paradigm proposed by \citep{acharya}. 
 Here, we show how to generalize this approach using power-divergence statistics. We can view goodness-of-fit hypothesis tests for a multinomial distribution as testing a hypothesis about the parameters $P = (\pi_1, \ldots, \pi_n)$, where each $\pi_i$ represents the probability of the $i$-th cell in the multinomial distribution. It is common to define the  null hypothesis $H_0: \Pi = \Pi_0$ where $\Pi_0 = (\pi_{01}, \ldots, \pi_{0n})$ is a pre-specified probability vector. The most widely-used test is Pearson's $\chi^2$ test, which gives the test statistic, given a IID sample of size $m$: 
\begin{equation}
    X^2 = \sum_{i=1}^n \frac{(X_i - m \pi_{0i})^2}{m \pi_{0i}}
\end{equation}
which is known to asymptotically have a $\chi^2$-distribution with $n-1$ degrees of freedom under $H_0$, and rejection of the hypothesis occurs when the observed value for $\chi^2$ is greater than equal to the pre-specified value found in the $\chi^2_{n-1}$ tables.

We now describe a more general paradigm for hypothesis testing for Bayes networks over multinomial distributions, using the {\em power divergence} goodness-of-fit tests proposed by \citep{cressie-read}. Under the assumption of sub-additivity, which holds for many $\phi$-divergences that satisfy an $\epsilon$-closeness property \citep{DBLP:conf/aistats/DingDF21}, we can use power divergence hypothesis tests as a more flexible and general paradigm than $\chi^2$-based tests for measuring causal influences. First, we introduce a directed divergence measure introduced by \citep{RATHIE197238} that is closely related to the power divergence test proposed by \citep{cressie-read}. 

\begin{definition}
Given two discrete probability distributions $p = (p_1, \ldots, p_n)$ and $q = (q_1, \ldots, q_n)$, the {\bf directed divergence} of type $\beta$ is defined as:
\begin{equation}
\label{dd}
    I^\beta_n(p \parallel q) = \frac{1}{2^{\beta - 1}} \left( \sum_{i=1}^n p_i^\beta q_i^{1 - \beta} - 1 \right), \ \ \mbox{for} \ \ \beta \neq 1
\end{equation}
\end{definition}

It can be easily shown that the directed divergence reduces to KL-divergence in the limit as $\beta$ tends to $1$, that is, $\lim_{\beta \rightarrow 1} I^\beta_n(p \parallel q) = KL(p \parallel q)$. We now introduce the power divergence between two discrete distributions $p$ and $q$ as follows \citep{cressie-read}: 

\begin{definition}
The {\bf power divergence} is defined as: 
\begin{equation}
\label{pd}
I^\lambda(p \parallel q) = \frac{1}{\lambda (\lambda + 1)} \sum_{i=1}^n p_i \left[ \left( \frac{p_i}{q_i}\right)^\lambda - 1 \right], \ \ \ -\infty < \lambda < \infty 
\end{equation}
\end{definition}

Since Equation~\ref{pd} is not defined for $\lambda = -1$ or $\lambda = 0$, for these two special cases, it is customary to define power divergences as the continuous limit as $\lim_{\lambda \rightarrow -1}$ and $\lim_{\lambda \rightarrow 0}$, respectively. Table~\ref{pd} shows how many common goodness-of-fit statistics emerge as special cases of the power divergence test. 

\begin{table}[t]
    \centering
    \begin{tabular}{|c|c|}  \hline
     {\bf Name}    &  $\lambda$ value  \\ \hline 
      $\chi^2$-test    &  $\lambda = 1$  \\ \hline 
     Neyman-modified $\chi^2$-test  &    $\lambda = -2$ \\ \hline 
       Loglikelihood ratio statistic  & $\lambda = -1$ \\ \hline 
        Freeman-Tukey statistic &  $\lambda = -\frac{1}{2}$ \\ \hline 
    \end{tabular}
    \caption{Power Divergence Test generalizes many common hypothesis tests.}
    \label{tab:my_label}
\end{table}

The directed divergence specified by Equation~\ref{dd} and the power divergence measure specified by Equation~\ref{pd} are related by the following identity \citep{cressie-read}: 
\begin{equation}
    I^\lambda(p \parallel q)   = \frac{2^\lambda - 1}{\lambda (\lambda + 1)} I^{\beta + 1}(p \parallel q), \ \ \ \lambda \neq -1 
\end{equation}
Given the original distribution $P = (p_1, \ldots, p_n)$ and the intervened distribution $P_S = (p^S_1, \ldots, p^S_n)$, we can formally define the {\em power-divergence} statistic for measuring causal influence as:

\begin{definition}
Given a causal model ${\cal M} = ({\cal G}, P)$, where $P$ is a multinomial distribution over $n$ variables $X_1, \ldots, X_n$, and $G = (V, E)$ is a DAG, the causal influence of intervening on a set of edges $S \subset E$ is defined by the power-divergence statistic as:
\begin{equation}
    D_\lambda(P \parallel P_S) =  \frac{1}{\lambda (\lambda + 1)} \sum_{i=1}^n p_i \left[ \left( \frac{p_i}{p^S_i}\right)^\lambda - 1 \right], \ \ \ -\infty < \lambda < \infty 
\end{equation}
\end{definition}

where as before, the special cases of $\lambda = -1$ and $\lambda = 0$ are treated using appropriate limits. We can show that the power divergence statistic is a special case of the $\phi$-divergence based causal influence model proposed in the paper by specifying the $\phi$ function as follows: 

\begin{definition}
The power-divergence statistic $D_\lambda(P \parallel P_S)$ is a special case of the $\phi$-divergence based causal influence model $D_\phi(P \parallel P_S)$, if we define $\phi$ as follows: 
\begin{equation}
    \phi(t) = \frac{t^{\lambda + 1} - t - \lambda (t - 1)}{\lambda (\lambda + 1)}, \ \ \ \lambda \neq 0, -1
\end{equation}
\end{definition}

Finally, we use the results shown by \citep{DBLP:conf/aistats/DingDF21} on sub-additivity of general $\phi$-divergences when the distributions $P$ and $P_S$ are ``close" to each other. 

\begin{definition}
We define the original distribution $P$ on a causal model ${\cal M} = (G, P)$, where $G = (V,E)$ is a DAG on $n$ discrete variables, as {\bf one-sided} $epsilon$-close to the intervened distribution $P_S$, for some subset $S \subset E$ and $0 < \epsilon < 1$, if for all $i \in [n], \frac{p_i}{p^S_i} < 1 + \epsilon$. Furthermore, $P$ and $P_S$ are {\bf two-sided} $\epsilon$-close if $1 - \epsilon < \frac{p_i}{p^S_i} < 1 + \epsilon$. 
\end{definition}

\citep{DBLP:conf/aistats/DingDF21} showed that most $\phi$-divergences are subadditive when they are applied to causal interventions that yield one-sided or two-sided $\epsilon$-close changes in the resulting distributions. 

\begin{theorem}
A $\phi$-divergence whose $\phi(t)$ function is continuous on $(0, \infty)$ and twice differentiable at $t = 1$ with $\phi^{``}(0) > 0$ satisfies $\alpha$ linear subadditivity when $P$ and $P_S$ are two-sided $\epsilon$-close with $\epsilon(\alpha) > 0$ where $\epsilon(\alpha)$ is a non-increasing function and $\lim_{\alpha \rightarrow 0} = 1$, where $\alpha$-linear subadditivity is defined as:
\begin{equation}
    D_\phi(P \parallel P_S) - \epsilon \leq \alpha \sum_{X_i  \in \mbox{trg}(S)} D_{\phi}(P_{X_i \cup \mbox{Pa}_i} \parallel P_{X_i \cup \mbox{Pa}^{\bar{S}}_i})
\end{equation}
\end{theorem}

where as before, given a set of intervened edges $S$, the target of intervention is $\mbox{trg}(S) = \{X_i | (X_j, X_i) \in S \}$. The proof of the above theorem follows directly from the corresponding result in \citep{DBLP:conf/aistats/DingDF21}. This theorem generalizes the causal influence decomposability for $\phi$-divergences given previously, assuming the interventions produce $\epsilon$-close changes.

\section{Causal Entropy of Structural Equation Models} 

Thus far, we have limited the discussion to discrete multinomial models. We extend the scope of causal entropy to linear structural equation models (SEMs) in this section, generalizing the treatment of SEMs developed in \citep{janzing}, which was restricted to the specific case of KL-divergence.  We follow the standard practice of assuming that there exists a total ordering of all variables $X_i, 1 \leq n$ in the model such that  each endogenous variable $X_i$ is a (deterministic) function of some subset of variables $X_j, 1 \leq j < i$.

\begin{definition}
The causal influence ${\cal C}_{X_i \rightarrow X_j}(P) = D_{KL}(P \parallel P_{X_i \rightarrow X_j})$ in a structural equation model (SEM) ${\cal M} = (X,\epsilon,f)$ compares the relative entropy between the pre-intervention distribution $P(x_1, \ldots, x_n)$, specified by set of deterministic equations $x_j = f_j(\mbox{pa}_j, \epsilon_j)$, where $\{\epsilon_1, \ldots, \epsilon_n \}$ represents a set of jointly independent unobserved noise variables, and the post-intervention distribution $P_{X_i \rightarrow X_j}$ defined by the modified SEM equations $x_l = f_l(\mbox{pa}_l, \epsilon_l)$, for $l \neq j$, and $x_j = f_j(\mbox{pa}_j \setminus \{x_j \}, (x'_i, \epsilon_j))$. 
\end{definition}
In effect, intervening on an edge $X_i \rightarrow X_j$ implies creating an IID copy of $X_i$, which is added to that to the noise term in the SEM definition of $X_j$. We can write the general SEM equation:
\begin{equation}
X_j = \sum_{(i,j) \in E} A_{ij} X_i + \epsilon_j
\end{equation}
where $E$ defines the set of comparable relations in the $k$-partite model (e.g., relations exist between layers, and not within layers). In vector form, we can write the SEM equation as:
\begin{equation}
X = A X + \epsilon, \ \ \ \mbox{or, equivalently} \ \ X = (I - A)^{-1} \epsilon
\end{equation}
where $A$ is a strictly lower triangular matrix with zero diagonals, and as before, the $\epsilon$ noise variables are jointly independent. In fact, $A$ is a block diagonal strictly lower triangular matrix to account for the $k$-partitioned structure. The covariance matrix of $X$ can be written as
\begin{equation}
\Sigma = (I - A)^{-1} \Sigma_\epsilon (I - A)^{-T}
\end{equation}
where $\Sigma_\epsilon$ is the covariance matrix of the noise variables $\epsilon$. Consider now an intervention on a subset $S$ of relations in the model, which we represent by decomposing $A = A_S + A_{\bar{S}}$, where $A_S$ comprises of relations in $S$ and $A_{\bar{S}}$ are those that are not in $S$. The modified SEM equations for the intervened system can be written as follows:
\begin{equation}
    X = A_{\bar{S}} X + \epsilon', \ \mbox{where} \ \epsilon' = \epsilon + A_{S} X'
\end{equation}
where $X' = (X_1, \ldots, X_n)$ is an IID copy of $X$, namely $X'_j$ is distributed like $X_j$, and all the variables $X', \epsilon$ are jointly independent. The covariance of the modified $\epsilon'$, and the intervened system $\Sigma_S$ is defined as: 
\begin{equation}
\Sigma_{\epsilon'} = \Sigma_\epsilon + A_S \Sigma^D_X A^T_S, \ \ \Sigma_S = (I - A_S)^{-1} \Sigma_E (I - A_S)^{-T}
\end{equation}
Assuming the noise variables $\epsilon$ are jointly independent, we can represent causal influence in terms of the $\phi$-divergence between two multivariate Gaussian distributions, where each distribution can be written as:
 \begin{equation}
     P(x | \mu, \Sigma) = \frac{1}{\sqrt{(2 \pi)^d | \Sigma |}}\exp{-\frac{1}{2}(x  - \mu)^T \Sigma^{-1} (x - \mu)}
 \end{equation}
 where $| \Sigma |$ denotes the determinant of the covariance matrix $\Sigma$, and $d$ is the dimension. 
 The squared Hellinger distance $D_{H^2}(P \parallel P_S)$ between the original distribution $P$ and the intervention distribution $P_S$, where $S \subset E$ is the set of edges intervened on, can be written as:
 \begin{equation}
     D_{H^2}(P \parallel P_S) = 1 - \frac{|\Sigma|^{\frac{1}{4}} |\Sigma_S|^{\frac{1}{4}}}{|\bar{\Sigma} | } \exp\{-\frac{1}{8} (\mu - \mu_S)^T \bar{\Sigma}^{-1} (\mu - \mu_S)\}
 \end{equation}
 where $\bar{\Sigma} = \frac{\Sigma + \Sigma_S}{2}$.  Similarly, the KL-divergence $D_{KL}(P \parallel P_S)$ between the original and intervened distribution is given as:
 \begin{equation}
     D_{KL}(P \parallel P_S) = \frac{1}{2}\left( \log \frac{|\Sigma_S|}{|\Sigma|} -d + \mbox{Tr}(\Sigma^{-1}_S \Sigma) + (\mu - \mu_S)^T \Sigma^{-1}_S (\mu - \mu_S) \right)
 \end{equation}
 If we assume that the variables $X_i$ are normalized so that they are mean $0$, then the causal influences simply as shown below, for the above two cases:
 \begin{equation}
     D_{H^2}(P \parallel P_S) = 1 - \frac{|\Sigma|^{\frac{1}{4}} |\Sigma_S|^{\frac{1}{4}}}{|\bar{\Sigma} | }
 \end{equation}
 \begin{equation}
     D_{KL}(P \parallel P_S) = \frac{1}{2}\left( \log \frac{|\Sigma_S|}{|\Sigma|} - d + \mbox{Tr}(\Sigma^{-1}_S \Sigma) \right)
 \end{equation}
 Finally, we can define the causal influence in $k$-partite SEM design, using the above definitions, again simplified for the mean $0$ centered case as:
\begin{definition}
\label{bcp1}
The {\bf $k$-partite average causal influence} $ACI^{H^2}(S,k)$ using squared Hellinger distance decomposes along each level: 
\begin{equation}
    ACI^{H^2}(S,k) = \frac{1}{|S|} \sum_{j=2}^k \sum_{i=1}^{|X_i|} {\cal C}_{S^j_i}(P)  = \sum_{j=2}^k \sum_{i=1}^{|X_i|}  \left(1 -  \frac{|\Sigma|^{\frac{1}{4}} |\Sigma_{S^j_i}|^{\frac{1}{4}}}{|\bar{\Sigma} | }\right)
\end{equation}
where $\bar{\Sigma} = \frac{\Sigma + \Sigma_{S^J_i}}{2}$. 
\end{definition}
\begin{definition}
\label{bcp2}
The {\bf $k$-partite average causal influence} using KL-divergence $ACI^{KL}(S,k)$ decomposes along each level: 
\begin{equation}
    ACI^{KL}(S,k) = \frac{1}{|S|} \sum_{j=2}^k \sum_{i=1}^{|X_i|} {\cal C}_{S^j_i}(P)  = \sum_{j=2}^k \sum_{i=1}^{|X_i|} \frac{1}{2} \left( \Tr (\Sigma^{-1}_{S^j_i} \Sigma) - \log \frac{\det \Sigma}{\det \Sigma_{S^j_i}} - d \right)
\end{equation}
\end{definition}

\section{Measuring Average Treatment Effect using Run Estimators} 

The fundamental problem in any causal experiment is to measure the effect of interventions. In bipartite design problems, such as estimating the effect of interventions on power plants on the cardiovascular health outcomes of the residents in nearby locations \citep{zigler2018bipartite}, the challenge is to design a  suitable average treatment effect estimator. In this section, we introduce a new ATE estimator based on the classic Wald Wolfowitz estimator based on {\em run statistics} \citep{noether},  which we define as the WW-ATE estimator. The advantage of our WW-ATE estimator for bipartite experiments, compared to previous estimators, such as correlational clustering \citep{DBLP:conf/nips/Pouget-AbadieAS19} or its recently proposed generalization, the exposure reweighted linear (ERL) estimator \citep{harshaw2021design}, is that it does not assume knowledge of the interference topology or assume linearity of the exposure or response functions. 

We give an intuitive characterization of WW-ATE run statistics estimator, before giving the formal definition of the WW-ATE estimator. \footnote{Francis Galton designed the first run estimator in 1876 for a dataset provided to him by Charles Darwin!}  Let us take the example of the exposure of residents to pollution from nearby power plants studied by \citep{zigler2018bipartite}. Let us assume that the outcome units, i.e. residents, have real-valued health outcomes $Y_1({\bf z}), Y_2({\bf z}), \ldots, Y_n({\bf z})$ in response to a particular treatment ${\bf z}$, namely interventions on nearby power plants by implementing some pollution control devices. In measuring average treatment effect, we want to compare the $n$ outcomes under one treatment ${\bf z_i}$ with the $n$ outcomes under a different treatment ${\bf z_j}$. We assume in this case that the potential outcome functions and exposure functions are {\em monotonic} functions of the treatment vectors, which generalizes the linear exposure-response model proposed in \citep{harshaw2021design}, where it is assumed both of these functions are linear. Each pair of interventions ${\bf z}_i$ and ${\bf z}_j$ yield a sequence of $2n$ potential outcomes:
\[ r_{ij} = (Y_1({\bf z}_i), \ldots, Y_n({\bf z}_i), Y_1({\bf z}_j), \ldots, Y_n({\bf z}_j) ) \]
Here, $r_{ij}$ is the run sample of potential outcomes based on the pair of tests ${\bf z}_i$ and ${\bf z}_j$, whose difference we seek to determine. For example, \citep{harshaw2021design} propose comparing the ATE between the test where all intervention units are treated, namely ${\bf z}_i = (+1, \ldots, +1)$, vs the test where none of the intervention units are treated, namely ${\bf z}_j = (-1, \ldots, -1)$. Let $r^s_{ij}$ be the sorted sequence of potential outcomes in descending order:
\[ r^s_{ij} = (Y_{p_1}({\bf z}_{u_1}), \ldots, Y_{p_{2n}}({\bf z}_{u_{2n}}) \]
where $u_k \in \{i, j\}$ and $Y_{p_l} > Y_{p_{l+1}}, 1 \leq l < 2n$. Intuitively, the idea behind the WW-ATE estimator is that if the intervention, e.g. pollution control at a power plant, is effective, the health outcomes measured for all nearby residents under the intervention should be higher than the health outcomes measured under no intervention. That is, all the potential outcomes $Y_1({\bf z}_i), \ldots, Y_n({\bf z}_i)$ should be larger than their corresponding values under no intervention, namely $Y_1({\bf z}_j), \ldots, Y_n({\bf z}_j)$. If the ordered sequence looks completely random with respect to whether an outcome unit was responding to intervention $i$ or $j$, then the causal effect of the intervention is not measurable. 

For each sorted run $r^s_{ij}$, define the symbol sequence $s_{ij}$ as
\[ s_{ij} = (s^{1}_{ij}, \ldots, s^{2n}_{ij}) \] 
where each symbol $s^l_{ij} \in \{+1, -1 \}$, where $s^l_{ij} = +1$ if the corresponding potential outcome in the $l$th spot in the ordered sequence represents the potential outcome under intervention $i$, and $s^l_{ij} = +1$ if the $l$th spot represents the potential outcome under control $j$. What WW-ATE measures is the run statistic, namely the number of contigous sequences of $+1$'s or $-1$'s to decide if the sequence is occurring randomly, or if there is an underlying pattern to the ordering that cannot be explained by chance. 
\begin{definition}
The {\bf Wald-Wolfowitz Average Treatment Effect Run Test} (WW-ATE) is defined over two sequences of real-valued observations $X_1, \ldots, X_n$ and $Y_1, \ldots, Y_n$ (representing the potential outcomes under treatment and control). It is assumed that the two sequences are generated independendently. 
\begin{itemize}
    \item The null hypothesis $H_0$: the two sequences $X_i$ and $Y_i$ come from identical distributions. 
    \item The alternate hypothesis $H_1$: The two sequences $X_i$ and $Y_i$ come from different distributions. 
    \item The WW-ATE test statistic is based on computing the number of runs $r$, after pooling the sequences into one sequence of length $2n$ and sorting the sequence in numerically descending order, and converting it to its symbol sequence. A run is a sequence of identical symbols (e.g., +1's or -1's). 
    \item The run test statistic $r$ can be shown to be asymptotically normally distributed, and hence its large sample test statistic is given by 
    \begin{equation}
        z = \frac{(r - \mathbb{E}(r))}{\mbox{Var}(r)} 
    \end{equation}
    \item The expected value and variance of $r$ is given by:
    \begin{equation}
        \mathbb{E}(r) = \frac{2 u v}{v + v} + 1, \ \ \ \ \mbox{Var(r)} = \frac{2 u v (2 u v - v - u))}{(v + u)^2 (u + v - 1)}
    \end{equation}
    where $u$ is the number of $+1$'s in the symbol sequence and $v$ is the number of $-1$'s in the symbol sequence. 
    \item To determine whether the null hypothesis $H_0$ should be rejected, and hence the intervention is causally identifiable, since the test statistic $z$ is asymptotically normal, we can compare its value with the standard Normal distribution for significance for a particular $p$-value. 
\end{itemize}
\end{definition}
For comparison, the estimator proposed in \citep{harshaw2021design} for bipartite experiments is given as, where $Y_i({\bf z})$ and $x_i({\bf z})$ is the potential outcome and exposure of unit $i$ under treatment ${\bf z}$: 
\begin{definition}
The {\bf exposure reweighted linear (ERL) estimator} \citep{harshaw2021design} of the average treatment effect is given as:
\begin{equation}
\label{erl}
\hat{\tau} = \frac{2}{n} \sum_{i=1}^n Y_i({\bf z}) \left( \frac{x_i({\bf z}) - \mathbb{E}[x_i({\bf z})]}{\mbox{Var}(x_i({\bf z}))} \right)
\end{equation}
\end{definition}
A detailed experimental study of the run estimator is being planned as part of our future work, which will give deeper insight into its efficacy for bipartite experiments, and its generalization to $k$-partite experiments.

\newpage 
%\bibliographystyle{plainnat}  
%\bibliography{references,allcitations}

\begin{thebibliography}{47}
\providecommand{\natexlab}[1]{#1}
\providecommand{\url}[1]{\texttt{#1}}
\expandafter\ifx\csname urlstyle\endcsname\relax
  \providecommand{\doi}[1]{doi: #1}\else
  \providecommand{\doi}{doi: \begingroup \urlstyle{rm}\Url}\fi

\bibitem[Acharya et~al.(2015)Acharya, Daskalakis, and
  Kamath]{DBLP:conf/nips/AcharyaDK15}
Jayadev Acharya, Constantinos Daskalakis, and Gautam Kamath.
\newblock Optimal testing for properties of distributions.
\newblock In Corinna Cortes, Neil~D. Lawrence, Daniel~D. Lee, Masashi Sugiyama,
  and Roman Garnett, editors, \emph{Advances in Neural Information Processing
  Systems 28: Annual Conference on Neural Information Processing Systems 2015,
  December 7-12, 2015, Montreal, Quebec, Canada}, pages 3591--3599, 2015.
\newblock URL
  \url{https://proceedings.neurips.cc/paper/2015/hash/1f36c15d6a3d18d52e8d493bc8187cb9-Abstract.html}.

\bibitem[Acharya et~al.(2018)Acharya, Bhattacharyya, Daskalakis, and
  Kandasamy]{acharya}
Jayadev Acharya, Arnab Bhattacharyya, Constantinos Daskalakis, and Saravanan
  Kandasamy.
\newblock Learning and testing causal models with interventions.
\newblock In S.~Bengio, H.~Wallach, H.~Larochelle, K.~Grauman, N.~Cesa-Bianchi,
  and R.~Garnett, editors, \emph{Advances in Neural Information Processing
  Systems}, volume~31. Curran Associates, Inc., 2018.
\newblock URL
  \url{https://proceedings.neurips.cc/paper/2018/file/78631a4bb5303be54fa1cfdcb958c00a-Paper.pdf}.

\bibitem[Ali and Silvey(1966)]{ali-silvey}
S.~M. Ali and S.~D. Silvey.
\newblock A general class of coefficients of divergence of one distribution
  from another.
\newblock \emph{Journal of the Royal Statistical Society: Series B
  (Methodological)}, 28\penalty0 (1):\penalty0 131--142, 1966.
\newblock \doi{https://doi.org/10.1111/j.2517-6161.1966.tb00626.x}.
\newblock URL
  \url{https://rss.onlinelibrary.wiley.com/doi/abs/10.1111/j.2517-6161.1966.tb00626.x}.

\bibitem[Ay and Polani(2008)]{nihat}
Nihat Ay and Daniel Polani.
\newblock Information flows in causal networks.
\newblock \emph{Adv. Complex Syst.}, 11\penalty0 (1):\penalty0 17--41, 2008.
\newblock \doi{10.1142/S0219525908001465}.
\newblock URL \url{https://doi.org/10.1142/S0219525908001465}.

\bibitem[Bailey et~al.(2007)Bailey, Dalmau, and
  Kolaitis]{DBLP:journals/dam/BaileyDK07}
Delbert~D. Bailey, V{\'{\i}}ctor Dalmau, and Phokion~G. Kolaitis.
\newblock Phase transitions of pp-complete satisfiability problems.
\newblock \emph{Discret. Appl. Math.}, 155\penalty0 (12):\penalty0 1627--1639,
  2007.
\newblock \doi{10.1016/j.dam.2006.09.014}.
\newblock URL \url{https://doi.org/10.1016/j.dam.2006.09.014}.

\bibitem[Brinkmann and McKay(2002)]{brinkmann}
Gunnar Brinkmann and Brendan~D. McKay.
\newblock Posets on up to 16 points.
\newblock \emph{Order}, 19:\penalty0 147‚Äì179, 2002.

\bibitem[Charles et~al.(2010)Charles, Chickering, Devanur, Jain, and
  Sanghi]{DBLP:conf/sigecom/CharlesCDJS10}
Denis~Xavier Charles, Max Chickering, Nikhil~R. Devanur, Kamal Jain, and Manan
  Sanghi.
\newblock Fast algorithms for finding matchings in lopsided bipartite graphs
  with applications to display ads.
\newblock In David~C. Parkes, Chrysanthos Dellarocas, and Moshe Tennenholtz,
  editors, \emph{Proceedings 11th {ACM} Conference on Electronic Commerce
  (EC-2010), Cambridge, Massachusetts, USA, June 7-11, 2010}, pages 121--128.
  {ACM}, 2010.
\newblock \doi{10.1145/1807342.1807362}.
\newblock URL \url{https://doi.org/10.1145/1807342.1807362}.

\bibitem[Cressie and Read(1984)]{cressie-read}
Noel Cressie and Timothy R.~C. Read.
\newblock Multinomial goodness-of-fit tests.
\newblock \emph{Journal of the Royal Statistical Society. Series B
  (Methodological)}, 46\penalty0 (3):\penalty0 440--464, 1984.
\newblock ISSN 00359246.
\newblock URL \url{http://www.jstor.org/stable/2345686}.

\bibitem[CSISZ\'AR(1967)]{csiszar}
I.~CSISZ\'AR.
\newblock Information-type measures of difference of probability distributions
  and indirect observation.
\newblock \emph{Studia Scientiarum Mathematicarum Hungarica}, 2:\penalty0
  229--318, 1967.
\newblock URL \url{https://ci.nii.ac.jp/naid/10028997448/en/}.

\bibitem[Daskalakis and Pan(2017)]{pmlr-v65-daskalakis17a}
Constantinos Daskalakis and Qinxuan Pan.
\newblock Square {Hellinger} subadditivity for {Bayesian} networks and its
  applications to identity testing.
\newblock In Satyen Kale and Ohad Shamir, editors, \emph{Proceedings of the
  2017 Conference on Learning Theory}, volume~65 of \emph{Proceedings of
  Machine Learning Research}, pages 697--703. PMLR, 07--10 Jul 2017.
\newblock URL \url{http://proceedings.mlr.press/v65/daskalakis17a.html}.

\bibitem[Dhar(1978)]{dhar:1978}
Deepak Dhar.
\newblock Entropy and phase transitions in partially ordered sets.
\newblock \emph{Journal of Mathematical Physics}, 19\penalty0 (8):\penalty0
  1711--1713, 1978.

\bibitem[Dhar(1980)]{dhar:1980}
Deepak Dhar.
\newblock Asymptotic enumeration of partially ordered sets.
\newblock \emph{Pacific Journal of Mathematics}, 90\penalty0 (2):\penalty0
  299--305, 1980.

\bibitem[Ding et~al.(2021)Ding, Daskalakis, and
  Feizi]{DBLP:conf/aistats/DingDF21}
Mucong Ding, Constantinos Daskalakis, and Soheil Feizi.
\newblock Gans with conditional independence graphs: On subadditivity of
  probability divergences.
\newblock In Arindam Banerjee and Kenji Fukumizu, editors, \emph{The 24th
  International Conference on Artificial Intelligence and Statistics, {AISTATS}
  2021, April 13-15, 2021, Virtual Event}, volume 130 of \emph{Proceedings of
  Machine Learning Research}, pages 3709--3717. {PMLR}, 2021.
\newblock URL \url{http://proceedings.mlr.press/v130/ding21e.html}.

\bibitem[Eberhardt(2008)]{DBLP:conf/uai/Eberhardt08}
Frederick Eberhardt.
\newblock Almost optimal intervention sets for causal discovery.
\newblock In David~A. McAllester and Petri Myllym{\"{a}}ki, editors,
  \emph{{UAI} 2008, Proceedings of the 24th Conference in Uncertainty in
  Artificial Intelligence, Helsinki, Finland, July 9-12, 2008}, pages 161--168.
  {AUAI} Press, 2008.
\newblock URL
  \url{https://dslpitt.org/uai/displayArticleDetails.jsp?mmnu=1\&smnu=2\&article\_id=1948\&proceeding\_id=24}.

\bibitem[Frieze and Tkocz(2020)]{DBLP:journals/siamdm/FriezeT20}
Alan~M. Frieze and Tomasz Tkocz.
\newblock Random graphs with a fixed maximum degree.
\newblock \emph{{SIAM} J. Discret. Math.}, 34\penalty0 (1):\penalty0 53--61,
  2020.
\newblock \doi{10.1137/19M1249928}.
\newblock URL \url{https://doi.org/10.1137/19M1249928}.

\bibitem[Harshaw et~al.(2021)Harshaw, Eisenstat, Mirrokni, and
  Pouget-Abadie]{harshaw2021design}
Christopher Harshaw, David Eisenstat, Vahab Mirrokni, and Jean Pouget-Abadie.
\newblock Design and analysis of bipartite experiments under a linear
  exposure-response model.
\newblock \emph{Arxiv}, 2021.

\bibitem[Hauser and B{\"{u}}hlmann(2012)]{DBLP:journals/jmlr/HauserB12}
Alain Hauser and Peter B{\"{u}}hlmann.
\newblock Characterization and greedy learning of interventional markov
  equivalence classes of directed acyclic graphs.
\newblock \emph{J. Mach. Learn. Res.}, 13:\penalty0 2409--2464, 2012.
\newblock URL \url{http://dl.acm.org/citation.cfm?id=2503320}.

\bibitem[Imbens and Rubin(2015)]{rubin-book}
Guido~W. Imbens and Donald~B. Rubin.
\newblock \emph{Causal Inference for Statistics, Social, and Biomedical
  Sciences: An Introduction}.
\newblock Cambridge University Press, USA, 2015.
\newblock ISBN 0521885884.

\bibitem[Jacot et~al.(2018)Jacot, Hongler, and
  Gabriel]{DBLP:conf/nips/JacotHG18}
Arthur Jacot, Cl{\'{e}}ment Hongler, and Franck Gabriel.
\newblock Neural tangent kernel: Convergence and generalization in neural
  networks.
\newblock In Samy Bengio, Hanna~M. Wallach, Hugo Larochelle, Kristen Grauman,
  Nicol{\`{o}} Cesa{-}Bianchi, and Roman Garnett, editors, \emph{Advances in
  Neural Information Processing Systems 31: Annual Conference on Neural
  Information Processing Systems 2018, NeurIPS 2018, December 3-8, 2018,
  Montr{\'{e}}al, Canada}, pages 8580--8589, 2018.
\newblock URL
  \url{https://proceedings.neurips.cc/paper/2018/hash/5a4be1fa34e62bb8a6ec6b91d2462f5a-Abstract.html}.

\bibitem[Janzing et~al.(2013)Janzing, Balduzzi, Grosse-Wentrup, and
  Scholkopf]{janzing}
Dominik Janzing, David Balduzzi, Moritz Grosse-Wentrup, and Bernhard
  Scholkopf.
\newblock {Quantifying causal influences}.
\newblock \emph{The Annals of Statistics}, 41\penalty0 (5):\penalty0 2324 --
  2358, 2013.
\newblock \doi{10.1214/13-AOS1145}.
\newblock URL \url{https://doi.org/10.1214/13-AOS1145}.

\bibitem[Kleitman and Rothschild(1979)]{kleitman:1979}
D.~J. Kleitman and B.L. Rothschild.
\newblock A phase transition on partial orders.
\newblock \emph{Physica}, 96A:\penalty0 254--259, 1979.

\bibitem[Kleitman and Rothschild(2001)]{kleitman:1975}
D.~J. Kleitman and B.L. Rothschild.
\newblock Asymptotic enumeration of partial orders on a finite set.
\newblock \emph{Transactions of the American Mathematical Society},
  205:\penalty0 213--233, 2001.

\bibitem[Kocaoglu et~al.(2017)Kocaoglu, Shanmugam, and
  Bareinboim]{DBLP:conf/nips/KocaogluSB17}
Murat Kocaoglu, Karthikeyan Shanmugam, and Elias Bareinboim.
\newblock Experimental design for learning causal graphs with latent variables.
\newblock In Isabelle Guyon, Ulrike von Luxburg, Samy Bengio, Hanna~M. Wallach,
  Rob Fergus, S.~V.~N. Vishwanathan, and Roman Garnett, editors, \emph{Advances
  in Neural Information Processing Systems 30: Annual Conference on Neural
  Information Processing Systems 2017, December 4-9, 2017, Long Beach, CA,
  {USA}}, pages 7018--7028, 2017.
\newblock URL
  \url{https://proceedings.neurips.cc/paper/2017/hash/291d43c696d8c3704cdbe0a72ade5f6c-Abstract.html}.

\bibitem[Li et~al.(2020)Li, Fang, Zeng, Maag, Tong, and
  Zhang]{DBLP:journals/geoinformatica/LiFZMTZ20}
Yiming Li, Jingzhi Fang, Yuxiang Zeng, Balz Maag, Yongxin Tong, and Lingyu
  Zhang.
\newblock Two-sided online bipartite matching in spatial data: experiments and
  analysis.
\newblock \emph{GeoInformatica}, 24\penalty0 (1):\penalty0 175--198, 2020.
\newblock \doi{10.1007/s10707-019-00359-w}.
\newblock URL \url{https://doi.org/10.1007/s10707-019-00359-w}.

\bibitem[Mao-cheng(1984)]{MAOCHENG198415}
CAI Mao-cheng.
\newblock On separating systems of graphs.
\newblock \emph{Discrete Mathematics}, 49\penalty0 (1):\penalty0 15--20, 1984.
\newblock ISSN 0012-365X.
\newblock \doi{https://doi.org/10.1016/0012-365X(84)90146-8}.
\newblock URL
  \url{https://www.sciencedirect.com/science/article/pii/0012365X84901468}.

\bibitem[Massey and Massey(2005)]{DBLP:conf/isit/MasseyM05}
James~L. Massey and Peter~C. Massey.
\newblock Conservation of mutual and directed information.
\newblock In \emph{Proceedings of the 2005 {IEEE} International Symposium on
  Information Theory, {ISIT} 2005, Adelaide, South Australia, Australia, 4-9
  September 2005}, pages 157--158. {IEEE}, 2005.
\newblock \doi{10.1109/ISIT.2005.1523313}.
\newblock URL \url{https://doi.org/10.1109/ISIT.2005.1523313}.

\bibitem[McAuley et~al.(2015)McAuley, Targett, Shi, and van~den
  Hengel]{DBLP:conf/sigir/McAuleyTSH15}
Julian~J. McAuley, Christopher Targett, Qinfeng Shi, and Anton van~den Hengel.
\newblock Image-based recommendations on styles and substitutes.
\newblock In Ricardo Baeza{-}Yates, Mounia Lalmas, Alistair Moffat, and
  Berthier~A. Ribeiro{-}Neto, editors, \emph{Proceedings of the 38th
  International {ACM} {SIGIR} Conference on Research and Development in
  Information Retrieval, Santiago, Chile, August 9-13, 2015}, pages 43--52.
  {ACM}, 2015.
\newblock \doi{10.1145/2766462.2767755}.
\newblock URL \url{https://doi.org/10.1145/2766462.2767755}.

\bibitem[Mirsky(1971)]{mirksy:dilworth}
L.~Mirsky.
\newblock A dual of {Dilworth's} decomposition theorem.
\newblock \emph{The American Mathematical Monthly}, 78\penalty0 (8):\penalty0
  876--877, 1971.
\newblock \doi{10.1080/00029890.1971.11992886}.
\newblock URL \url{https://doi.org/10.1080/00029890.1971.11992886}.

\bibitem[Noether(1950)]{noether}
Gottfried~Emanuel Noether.
\newblock {Asymptotic Properties of the Wald-Wolfowitz Test of Randomness}.
\newblock \emph{The Annals of Mathematical Statistics}, 21\penalty0
  (2):\penalty0 231 -- 246, 1950.
\newblock \doi{10.1214/aoms/1177729841}.
\newblock URL \url{https://doi.org/10.1214/aoms/1177729841}.

\bibitem[Pearl(1989)]{pearl:bnets-book}
Judea Pearl.
\newblock \emph{Probabilistic reasoning in intelligent systems - networks of
  plausible inference}.
\newblock Morgan Kaufmann series in representation and reasoning. Morgan
  Kaufmann, 1989.

\bibitem[Pearl(2009)]{pearl:causalitybook}
Judea Pearl.
\newblock \emph{Causality: Models, Reasoning and Inference}.
\newblock Cambridge University Press, USA, 2nd edition, 2009.
\newblock ISBN 052189560X.

\bibitem[Pouget{-}Abadie et~al.(2019)Pouget{-}Abadie, Aydin, Schudy, Brodersen,
  and Mirrokni]{DBLP:conf/nips/Pouget-AbadieAS19}
Jean Pouget{-}Abadie, Kevin Aydin, Warren Schudy, Kay Brodersen, and Vahab~S.
  Mirrokni.
\newblock Variance reduction in bipartite experiments through correlation
  clustering.
\newblock In Hanna~M. Wallach, Hugo Larochelle, Alina Beygelzimer, Florence
  d'Alch{\'{e}}{-}Buc, Emily~B. Fox, and Roman Garnett, editors, \emph{Advances
  in Neural Information Processing Systems 32: Annual Conference on Neural
  Information Processing Systems 2019, NeurIPS 2019, December 8-14, 2019,
  Vancouver, BC, Canada}, pages 13288--13298, 2019.
\newblock URL
  \url{https://proceedings.neurips.cc/paper/2019/hash/bc047286b224b7bfa73d4cb02de1238d-Abstract.html}.

\bibitem[Pr\"{o}mel et~al.(2001{\natexlab{a}})Pr\"{o}mel, Steger, and
  Taraz]{promel:2001}
Hans Pr\"{o}mel, Angelika Steger, and Anusch Taraz.
\newblock Asymptotic enumeration, global structure, and constrained evolution.
\newblock \emph{Discrete Mathematics}, 229:\penalty0 205--220,
  2001{\natexlab{a}}.

\bibitem[Pr\"{o}mel et~al.(2001{\natexlab{b}})Pr\"{o}mel, Steger, and
  Taraz]{promel:phase}
Hans Pr\"{o}mel, Angelika Steger, and Anusch Taraz.
\newblock Phase transitions in the evolution of partial orders.
\newblock \emph{Journal of Combinatorial Theory}, 94:\penalty0 230--275,
  2001{\natexlab{b}}.

\bibitem[Raginsky(2011)]{DBLP:conf/allerton/Raginsky11}
Maxim Raginsky.
\newblock Directed information and pearl's causal calculus.
\newblock In \emph{49th Annual Allerton Conference on Communication, Control,
  and Computing, Allerton 2011, Allerton Park {\&} Retreat Center, Monticello,
  IL, USA, 28-30 September, 2011}, pages 958--965. {IEEE}, 2011.
\newblock \doi{10.1109/Allerton.2011.6120270}.
\newblock URL \url{https://doi.org/10.1109/Allerton.2011.6120270}.

\bibitem[Rathie and Kannappan(1972)]{RATHIE197238}
P.N. Rathie and Pl. Kannappan.
\newblock A directed-divergence function of type $\beta$.
\newblock \emph{Information and Control}, 20\penalty0 (1):\penalty0 38--45,
  1972.
\newblock ISSN 0019-9958.
\newblock \doi{https://doi.org/10.1016/S0019-9958(72)90260-4}.
\newblock URL
  \url{https://www.sciencedirect.com/science/article/pii/S0019995872902604}.

\bibitem[Robinson(1977)]{robinson}
R.~W. Robinson.
\newblock Counting unlabeled acyclic digraphs.
\newblock In Charles H.~C. Little, editor, \emph{Combinatorial Mathematics V},
  pages 28--43, Berlin, Heidelberg, 1977. Springer Berlin Heidelberg.
\newblock ISBN 978-3-540-37020-8.

\bibitem[R{\"{o}}dder et~al.(2019)R{\"{o}}dder, Dellnitz, Kulmann, Litzinger,
  and Reucher]{DBLP:journals/entropy/RodderDKLR19}
Wilhelm R{\"{o}}dder, Andreas Dellnitz, Friedhelm Kulmann, Sebastian Litzinger,
  and Elmar Reucher.
\newblock Bipartite structures in social networks: Traditional versus
  entropy-driven analyses.
\newblock \emph{Entropy}, 21\penalty0 (3):\penalty0 277, 2019.
\newblock \doi{10.3390/e21030277}.
\newblock URL \url{https://doi.org/10.3390/e21030277}.

\bibitem[Sason and Verd{\'{u}}(2015)]{DBLP:journals/corr/SasonV15}
Igal Sason and Sergio Verd{\'{u}}.
\newblock Bounds among {\textdollar}f{\textdollar}-divergences.
\newblock \emph{CoRR}, abs/1508.00335, 2015.
\newblock URL \url{http://arxiv.org/abs/1508.00335}.

\bibitem[Schlosser and Boissier(2018)]{DBLP:conf/kdd/Schlosser018}
Rainer Schlosser and Martin Boissier.
\newblock Dynamic pricing under competition on online marketplaces: {A}
  data-driven approach.
\newblock In Yike Guo and Faisal Farooq, editors, \emph{Proceedings of the 24th
  {ACM} {SIGKDD} International Conference on Knowledge Discovery {\&} Data
  Mining, {KDD} 2018, London, UK, August 19-23, 2018}, pages 705--714. {ACM},
  2018.
\newblock \doi{10.1145/3219819.3219833}.
\newblock URL \url{https://doi.org/10.1145/3219819.3219833}.

\bibitem[Schlosser et~al.(2018)Schlosser, Walther, Boissier, and
  Uflacker]{DBLP:conf/ijcai/SchlosserW0U18}
Rainer Schlosser, Carsten Walther, Martin Boissier, and Matthias Uflacker.
\newblock Data-driven inventory management and dynamic pricing competition on
  online marketplaces.
\newblock In J{\'{e}}r{\^{o}}me Lang, editor, \emph{Proceedings of the
  Twenty-Seventh International Joint Conference on Artificial Intelligence,
  {IJCAI} 2018, July 13-19, 2018, Stockholm, Sweden}, pages 5856--5858.
  ijcai.org, 2018.
\newblock \doi{10.24963/ijcai.2018/861}.
\newblock URL \url{https://doi.org/10.24963/ijcai.2018/861}.

\bibitem[Spirtes et~al.(2000)Spirtes, Glymour, and Scheines]{spirtes:book}
Peter Spirtes, Clark Glymour, and Richard Scheines.
\newblock \emph{Causation, Prediction, and Search, Second Edition}.
\newblock Adaptive computation and machine learning. {MIT} Press, 2000.
\newblock ISBN 978-0-262-19440-2.

\bibitem[Szemer\"{e}di(1975)]{szemeredi:1975}
E.~Szemer\"{e}di.
\newblock Regular partitions of graphs.
\newblock Technical Report STAN-CS-75-489, Stanford University, 1975.

\bibitem[Tadepalli and Russell(2021)]{prasad:aaai21}
Prasad Tadepalli and Stuart Russell.
\newblock {PAC} learning of causal trees with latent variables.
\newblock In \emph{AAAI}, 2021.

\bibitem[Taraz(1999)]{Taraz1999Phase}
Anuschirawan~Ralf Taraz.
\newblock \emph{Phase transitions in the evolution of partially ordered sets}.
\newblock PhD thesis, Humboldt-University zu Berlin,
  Mathematisch-Naturwissenschaftliche Fakulty II, 1999.

\bibitem[Wieczorek and Roth(2019)]{wieczorek}
Aleksander Wieczorek and Volker Roth.
\newblock Information theoretic causal effect quantification.
\newblock \emph{Entropy}, 21\penalty0 (10), 2019.
\newblock ISSN 1099-4300.
\newblock \doi{10.3390/e21100975}.
\newblock URL \url{https://www.mdpi.com/1099-4300/21/10/975}.

\bibitem[Zigler and Papadogeorgou(2018)]{zigler2018bipartite}
Corwin~M. Zigler and Georgia Papadogeorgou.
\newblock Bipartite causal inference with interference, 2018.

\end{thebibliography}

\end{document}